\pdfoutput=1

\documentclass[11pt]{article}

\usepackage[final]{acl}
\usepackage{times}
\usepackage{latexsym}
\usepackage[utf8]{inputenc}

\usepackage{graphicx}
\usepackage{microtype}
\usepackage{hyperref}
\usepackage{url}
\usepackage{booktabs}
\usepackage{multirow}
\usepackage{quoting}
\usepackage{makecell}

\usepackage{nicematrix}
\usepackage{tabularray}
\usepackage{longtable}
\usepackage{pifont}
\usepackage{tikz}

\usepackage{wrapfig}
\usepackage{float}
\usepackage{array}

\newcommand{\err}[1]{\textcolor{red}{#1}}

\usepackage{tcolorbox}
\usepackage{listings}
\usepackage[table]{xcolor} 
\usepackage{xcolor}
\usepackage{fancyvrb}
\definecolor{lightblue}{RGB}{46,110,187}
\definecolor{darkred}{RGB}{150,38,31}
\definecolor{darkgreen}{rgb}{0.0, 0.75, 0.0}
\definecolor{blue}{HTML}{3572EF}

\usepackage{lipsum}
\usepackage{enumitem}
\usepackage{lineno}
\usepackage[T2A,T1]{fontenc}
\usepackage[main=english]{babel}
\babelprovide{russian}

\usepackage{CJKutf8}

\definecolor{darkblue}{rgb}{0, 0, 0.5}
\hypersetup{colorlinks=true, citecolor=darkblue, linkcolor=darkblue, urlcolor=darkblue}

\title{Evaluating the Impact of Verbal Multiword Expressions on Machine Translation}

\author{Linfeng Liu, Saptarshi Ghosh, Tianyu Jiang \\
University of Cincinnati \\
\texttt{\{liu2lf, ghosh2si\}@mail.uc.edu, tianyu.jiang@uc.edu} \\
}

\begin{document}

\maketitle

\begin{abstract}

Verbal multiword expressions (VMWEs) remain difficult for machine translation because their meanings are often not recoverable from their component words. In this study, we analyze the impact of three VMWE categories---verbal idioms, verb-particle constructions, and light verb constructions---on machine translation quality from English to multiple languages. Using both established multiword expression datasets and standard machine translation datasets, we evaluate how state-of-the-art translation systems handle these expressions. Our experimental results consistently show that VMWEs negatively affect translation quality, with deeper analysis indicating that this degradation is primarily attributable to the VMWE itself rather than general sentence-level difficulty. We release our code and evaluation framework to test new MT systems for the community.\footnote{\url{https://github.com/cincynlp/vmwe-mt-eval}}

\end{abstract}

\section{Introduction}

\begin{figure}[t]
    \centering
    \includegraphics[width=\columnwidth]{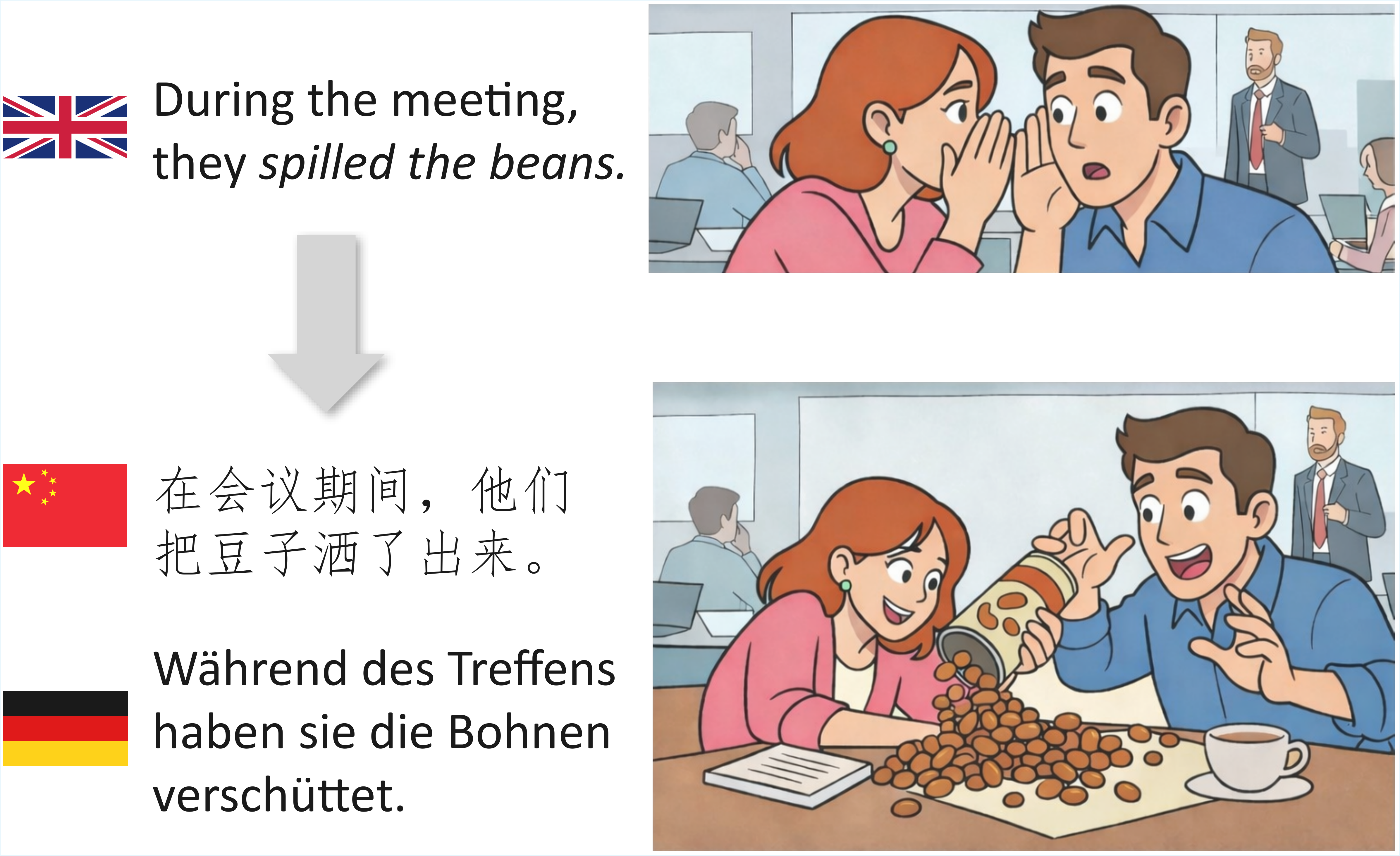}
    \caption{Illustrative example of incorrect machine translation of a sentence containing the verbal multiword expression \textit{spill the beans} (``reveal a secret''), where literal translations into Chinese and German lead to a semantically incorrect interpretation.}
    \label{fig:intro_img}
\end{figure}

Machine translation (MT) has achieved remarkable progress with the advent of neural architectures and large language models~\citep{lu-etal-2024-llamax, gemma2}, yet certain linguistic phenomena continue to challenge advanced systems, such as structural differences~\citep{wang2022progress} and morphologically complex words~\citep{weller-di-marco-fraser-2022-test}. \textit{Verbal multiword expressions} (VMWEs)---fixed or semi-fixed combinations of words containing a verb with non-compositional meanings---represent one of these persistent challenges. Unlike regular word combinations whose meanings derive from their constituent parts, multiword expressions require contextual understanding and often language-specific knowledge to translate accurately. For example, humans understand ``\textit{spill the beans}'' to mean \textit{reveal a secret}, but translation systems often render it literally, as shown in Figure~\ref{fig:intro_img}. Similarly, when people ``\textit{land on their feet}'' after a setback, or ``\textit{check in}'' at a hotel, the meanings emerge from the word combinations rather than individual components. Despite their prevalence in everyday language and their importance for meaning preservation in translation, VMWEs remain understudied in the context of MT systems~\citep{ramisch-etal-2013-hard,cap-etal-2015-account}. 

Verbal multiword expressions have long been recognized as an important challenge in NLP~\citep{Fellbaum2006, boukobza-rappoport-2009-multi,rohanian-etal-2020-verbal}. 
Following previous work on the most widely recognized verbal multiword expressions~\citep{constant-etal-2017-survey,savary-etal-2017-parseme}, we focus on three linguistically distinct VMWE types in English: (1) verbal idioms, whose meanings cannot be derived from component words (e.g., \textit{take the lion's share}), (2) verb-particle constructions, where particles modify verb meanings in non-trivial ways (e.g., \textit{figure out}), and (3) light verb constructions, where verbs primarily serve grammatical functions while nouns carry more semantic weight (e.g., \textit{give a speech}).

In this work, we present a comprehensive analysis of how these different categories of VMWEs affect translation quality across multiple state-of-the-art MT systems. Our investigation employs a novel dual-dataset approach, examining both established VMWE datasets (e.g., \citealp{saxena2020epie, Tu2012SortingOT}) and VMWE sentences extracted from the Conference on Machine Translation (WMT) datasets~\citep{kocmi-etal-2024-findings}. We evaluate translation quality through two methods: reference-free quality estimation~\citep{metricx, xcomet} and human direct assessment (DA) scores~\citep{graham-etal-2013-continuous}. We also release this evaluation pipeline as a reusable framework for testing additional MT systems under the same VMWE settings.

We conduct experiments on seven language pairs and eight state-of-the-art MT systems: four specific neural MT models, three LLM-based models, and Google Translate API. Our results demonstrate that verbal multiword expressions consistently degrade translation quality, with the degree of degradation correlating with non-compositionality. We perform deeper analysis and show that VMWE presence is the primary cause for translation degradation rather than confounding factors such as linguistic complexity. In summary, the key contributions are:
\begin{itemize}[itemsep=1pt, topsep=5pt]
    \item We provide the first comprehensive study of the impact of verbal multiword expressions on machine translation quality, and we release the evaluation framework for reproducible testing of new MT systems. 
    \item Our experimental results quantitatively demonstrate that VMWEs degrade translation quality across multiple MT systems and language pairs, with the degree of degradation correlating with the level of non-compositionality.
    \item We analyze VMWE-related translation difficulty with error-span analysis and regression controls, showing that the degradation persists after accounting for standard sentence-level difficulty factors and that errors often concentrate on the VMWE span.
\end{itemize}

\section{Related Work}

\textbf{Verbal Multiword Expressions in NLP.} Multiword expressions (MWEs) have long been recognized as important challenges in natural language processing due to their idiosyncratic, non-compositional, and often ambiguous nature~\citep{baldwin, baldwin2010multiword, constant-etal-2017-survey}. Among MWEs, verbal multiword expressions (VMWEs) such as light verb constructions, verb-particle constructions, and verbal idioms have drawn considerable attention~\citep{mwe_acquisitions, pasquer-etal-2020-verbal, rohanian-etal-2020-verbal,jiang-riloff-2022-identifying,parseme}. These observations have led to considerable work toward constructing annotations and datasets for individual VMWE types. For example, \citet{tu-roth-2011-learning} and \citet{Tu2012SortingOT} compiled datasets for English light verbs and verb-particles. \citet{saxena2020epie} and \citet{haagsma-etal-2020-magpie} introduced datasets of idiomatic expressions. VMWEs have also been widely studied for their cross-linguistic variability, as languages differ both in the lexicon and syntactic patterns employed to express them~\citep{syntax_based_lvc, racz-etal-2014-4fx, multilingualidioms}. While English uses light verbs like \textit{make} and \textit{give}, other languages may use unique verbs not equivalent in English for similar constructions~\citep{Butt_2010}. Other studies have explored idiomatic expressions of low resource languages~\citep{dimakis-etal-2024-dictionary, ohuoba-etal-2024-quantifying}. Annotation efforts like the \textsc{parseme} shared task~\citep{savary-etal-2017-parseme, parseme} have standardized VMWE identification across languages. 

\noindent \textbf{MWE in Machine Translation.} Machine Translation (MT) has seen significant progress with the transition from phrase-based statistical MT~\citep{stat_based_translation, koehn-knowles-2017-six} to neural MT~\citep{Bahdanau2014NeuralMT, attention}. More recently, large language models (LLMs) have been applied to MT tasks, bringing improvement in fluency and contextual understanding~\citep{manakhimova-etal-2023-linguistically, kocmi-etal-2024-findings}. Researchers have known for a long time that MWEs are essential for MT systems. Earlier work by~\citet{carpuat-diab-2010-task} examined idiom translation in statistical MT and highlighted the mismatch between compositional meaning and phrase-based alignment. With the advent of neural MT, numerous studies have focused on improving MWE translation with neural models~\citep{koehn-knowles-2017-six, zaninello-birch-2020-multiword, baziotis-etal-2023-automatic}. More recently, 
\citet{song-xu-2024-deep} analyzed the effects of MWEs and named entities in Chinese–English MT, finding that errors involving idioms significantly affect overall quality. However, despite these advances, there remains a notable lack of systematic evaluation of VMWEs in state-of-the-art machine translation systems. Our work provides the first comprehensive study of the impact of verbal multiword expressions (VMWEs) on machine translation (MT) quality.

\section{Evaluating Translation on VMWE Datasets}
\label{section_2}

Having reviewed the linguistic background and prior work, we now turn to our evaluation setup. This section describes the VMWE categories we study, the datasets and filtering procedures, the MT systems we evaluate, and the metrics used to assess translation quality. We translate the source sentences with each MT system and score the outputs using reference-free, quality estimation models.


\subsection{Verbal Multiword Expression Types}

Following previous research~\citep{constant-etal-2017-survey, savary-etal-2017-parseme}, we focus on three types of common verbal multiword expressions in English: verbal idioms, verb-particle constructions and light verb constructions. These categories challenge natural language processing due to their semantic opacity, context dependence and idiomaticity. 

\noindent \textbf{Verbal idiom (VID)}: A fixed phrase with a verb whose meaning cannot be guessed from the individual words (e.g., \textit{spill the beans} means to \textit{reveal the secret}). These expressions are highly non-compositional and often require contextual or cultural understanding.

\noindent \textbf{Verb-Particle Construction (VPC)}: Consists of a verb and a particle (typically an adverb or preposition) that together form a new meaning (e.g., \textit{give up} means \textit{quit}). These expressions are generally semi-compositional, as the meaning is not fully predictable from the parts.

\noindent \textbf{Light Verb Construction (LVC)}: Combines a verb with little meaning on its own (e.g., \textit{take}) and a predicative noun (e.g., \textit{walk}) to form a full expression (e.g., \textit{take a walk}). These constructions are semi-compositional, with the noun carrying most of the meaning.

\subsection{Datasets}
\label{existing_datasets}

Verbal multiword expressions have long been recognized as a challenge in natural language processing. Therefore, there have been multiple corpora dedicated to VMWE and its subcategories.

\noindent \textbf{Verbal Idiom datasets.} We use these two datasets: 1) EPIE~\citep{saxena2020epie}: This dataset contains annotated formal idiomatic English sentences extracted from the British National Corpus (BNC)~\citep{bnc1994xml}. 2) MAGPIE~\citep{haagsma-etal-2020-magpie}: A large sense-annotated corpus based on the BNC and Parallel Meaning Bank (PMB)~\citep{abzianidze-etal-2017-parallel}. We retain only idiom sentences from these two corpora that include a verb. This filtering step ensures our dataset accurately reflects verbal idiomatic expressions. After filtering, we sample 2,000 sentences, 1,000 from each corpus. 

\noindent \textbf{Verb-Particle Construction datasets.} We use the ~\citet{Tu2012SortingOT} VPC dataset consisting of 1,348 sentences sampled from BNC dataset, of which 878 sentences with true verb-particle construction and 470 sentences with simplex verb-preposition combination sentences, focusing on combinations of six verbs \textit{do}, \textit{get}, \textit{give}, \textit{have}, \textit{make} and \textit{take} with nineteen common prepositions or particles. We only keep the 878 sentences that contain the true verb-particle construction.

\noindent \textbf{Light Verb Construction datasets.} We use the ~\citet{tu-roth-2011-learning} LVC dataset. It is a balanced, annotated dataset of English light verb constructions using the BNC, includes 1,039 sentences that genuinely contain light verb constructions and 1,123 simplex verb-noun sentences. This dataset focuses on the same six common English light verbs used in the VPC dataset. We used all 1,039 true light verb construction sentences for evaluation. 

\noindent \textbf{Non-VMWE Sentences.} To accurately assess the impact of VMWEs on translation quality, we constructed a comparison dataset consisting of sentences without verbal multiword expressions. We obtained these non-VMWE sentences by applying rule-based heuristics to filter the BNC. Verbal idioms are excluded using idiom dictionaries from the EPIE and MAGPIE corpora. Verb-particle constructions are filtered using the spaCy dependency parser~\citep{spacy2020} where a particle has a \textit{prt} relation to a verb. Light verb constructions are detected using the light verb lists provided by \citet{Tu2012SortingOT} and \citet{Huddleston_Huddleston_Pullum_2002}. Finally we randomly sampled 2,000 sentences as our non-VMWE dataset from the filtered result.

\subsection{QE Systems \& MT Models Used}


For machine translation quality assessment, we utilize two state-of-the-art quality estimation (QE) models from the WMT24 Metrics Shared Task~\citep{freitag-etal-2024-llms}. Unlike most evaluation models, these two models can evaluate a translation without a reference. They only need the original sentence and the machine translation to assess quality.

\noindent \textbf{1) MetricX-24-QE}~\citep{metricx}: Based on mT5~\citep{xue-etal-2021-mt5}, it is fine-tuned on a combination of human evaluation scores and synthetic data designed to capture common translation errors. In QE mode, it generates a score in the range from 0 to 25, where lower scores indicate better translation quality.

\noindent \textbf{2) xCOMET-QE}~\citep{xcomet}: Built on a large pre-trained encoder model XLM-RoBERTa-XL~\citep{goyal-etal-2021-larger}, it is trained using a multi-task setup that combines sentence-level quality assessments and word-level severity tags, in order to predict both a sentence-level score and identify specific error spans with their severity levels. In QE mode, it produces a score between 0 and 1, with higher scores indicating better translation quality.

\noindent \textbf{MT Models:} We used 8 MT systems in this work: 1) SeamlessM4T~\citep{seamless2023}, 2) Madlad400~\citep{kudugunta2023madlad400}, 3) M2M100~\citep{fan2020englishcentric}, 4) Opus-MT~\citep{tiedemann-thottingal-2020-opus}, 5) LLaMAX3 Alpaca~\citep{lu-etal-2024-llamax}, 6) Phi-4-multimodal~\citep{microsoft2025phi4minitechnicalreportcompact}, 7) GemmaX2~\citep{cui2025multilingualmachinetranslationopen}, 8) Google Translate API. See further details about these MT systems in Appendix~\ref{mt_systems}. 

\noindent \textbf{Language scope.} For VMWE-dataset experiments, we translate from English into seven targets: German (de), Czech (cs), Russian (ru), Chinese (zh), Spanish (es), Japanese (ja), and Turkish (tr). Selected to span major language families and to be supported by our QE models and MT systems.

\section{Evaluating Translation on WMT Datasets}

\label{section_3}
The experiment design above uses reference-free evaluation on VMWE-specific datasets. To test whether the same pattern holds under human evaluation, and to strengthen external validity, we turn to datasets from the Conference on Machine Translation (WMT), which provide outputs from multiple machine translation systems along with human judgments in the form of direct assessment (DA) scores~\citep{graham-etal-2013-continuous}. Details about human DA scores are given in Appendix~\ref{app:da}.

In the WMT setting, we (i) locate VMWE instances in English source sentences, (ii) collect the corresponding system translations and human DA scores, and (iii) measure the difference in human-rated quality between VMWE and non-VMWE subsets. This lets us ask the same question as before—do VMWEs hurt translation quality?—but now answer with human judgments.

\subsection{Dataset}
\label{wmt_dataset}

The Conference on Machine Translation (WMT) datasets are widely used for benchmarking MT models. The datasets across multiple years contain English sentences along with their corresponding translations in other languages by MT systems and humans, evaluated by humans ~\citep{bojar-etal-2016-findings}.

To maintain a balance between diversity and consistency, we select the four most frequently used language pairs from 2017 to 2024: English to Czech (\textit{en-cs}), English to German (\textit{en-de}), English to Chinese (\textit{en-zh}) and English to Russian (\textit{en-ru}).

Different MT systems are evaluated and reported by the WMT Findings of each year---\cite{bojar-etal-2017-findings}, \cite{bojar-etal-2018-findings}, \cite{barrault-etal-2019-findings}, \cite{barrault-etal-2020-findings}, \cite{akhbardeh-etal-2021-findings}, \cite{kocmi-etal-2022-findings}, \cite{kocmi-etal-2023-findings}, \cite{kocmi-etal-2024-findings}---from 2017 to 2024. We selected the top four MT systems for each year's specific language pair to reflect the highest level of MT performance in that year.


\begin{figure*}[ht]
    \centering
    \includegraphics[width=0.92\linewidth]{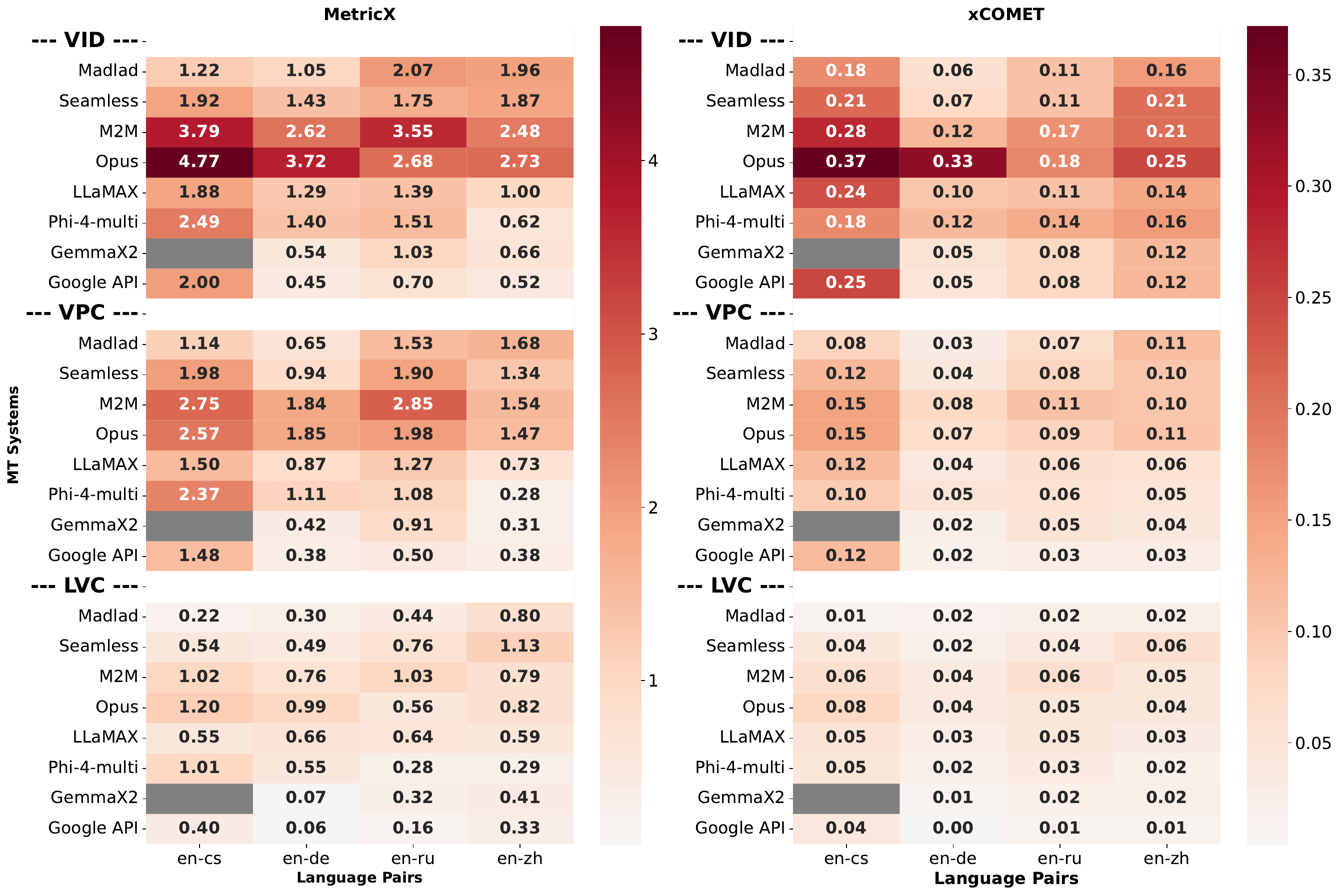}
    \caption{Comparison of the translation quality between sentences with and without VMWE, using QE models MetricX-24 (\textit{range from 0 to 25, lower the better}) and xCOMET (\textit{range from 0 to 1, higher the better}) scores. For each cell, the values are the difference between the score of VMWE and non-VMWE sentences. 
    Larger numbers (darker colors) mean worse translation performance on VMWE sentences.}
    \label{fig:heatmap_metric}
\end{figure*}

\subsection{VMWE Extraction and Classification Using Large Language Models (LLMs)}
\label{sec:vmwe_extract}

We construct VMWE and non-VMWE subsets from the WMT data as follows:
For non-VMWE sentences, we apply the same procedure as in Section~\ref{existing_datasets} to the WMT datasets.
To obtain VMWE sentences, we apply a two-step extraction pipeline:

\noindent\textbf{1) Heuristic retrieval.} For verbal idioms: Starting from filtered idiom lists in EPIE and MAGPIE, we retrieve sentences containing similar phrases, requiring a BLEU-4 score of at least 0.6~\citep{papineni-etal-2002-bleu}. We use dependency parser for potential verb-particles and light verbs.

\noindent\textbf{2) LLM-based disambiguation.} Following the \textsc{parseme} guidelines~\citep{parseme}, we design a chain-of-thought prompt~\citep{chain_of_thought} and since there are no gold labels in WMT to compute accuracy, we test the performance of the LLMs on the existing VMWE datasets. GPT-4o~\citep{openai2024gpt4ocard} is the best performing model, so we use it for extracting VMWE sentences from WMT. The performance of GPT-4o is given in Table~\ref{table:llm_performance}, and the results for all LLMs and full pipeline details are provided in Appendix~\ref{app:llm_comparison_sec}. Finally, we compare translation quality between the VMWEs and non-VMWEs using their respective human DA scores.

\begin{table}[t]
    \centering
    \resizebox{0.6\linewidth}{!}{
    \begin{tabular}{lccc}
    \toprule
    \textbf{Type} & \textbf{VID} & \textbf{VPC} & \textbf{LVC} \\
    \midrule
    Pos & 81.8 & 80.0 & 81.6\\
    Neg & 78.9 & 84.4 & 78.0\\
    \bottomrule
    \end{tabular}
    }
\caption{The F1-score of the GPT-4o model's performance for identifying VMWE sentences on a specific balanced sampled dataset. Pos: VMWE sentences. Neg: non-VMWE sentences.}
\label{table:llm_performance}
\end{table}

\section{Results}

We report the translation quality comparing sentences with and without verbal multiword expressions on both the specific VMWE datasets and the machine translation datasets (WMT). Our experimental results suggest that VMWEs consistently degrade machine translation quality.

\begin{figure*}[t]
    \centering
    \includegraphics[width=0.9\linewidth]{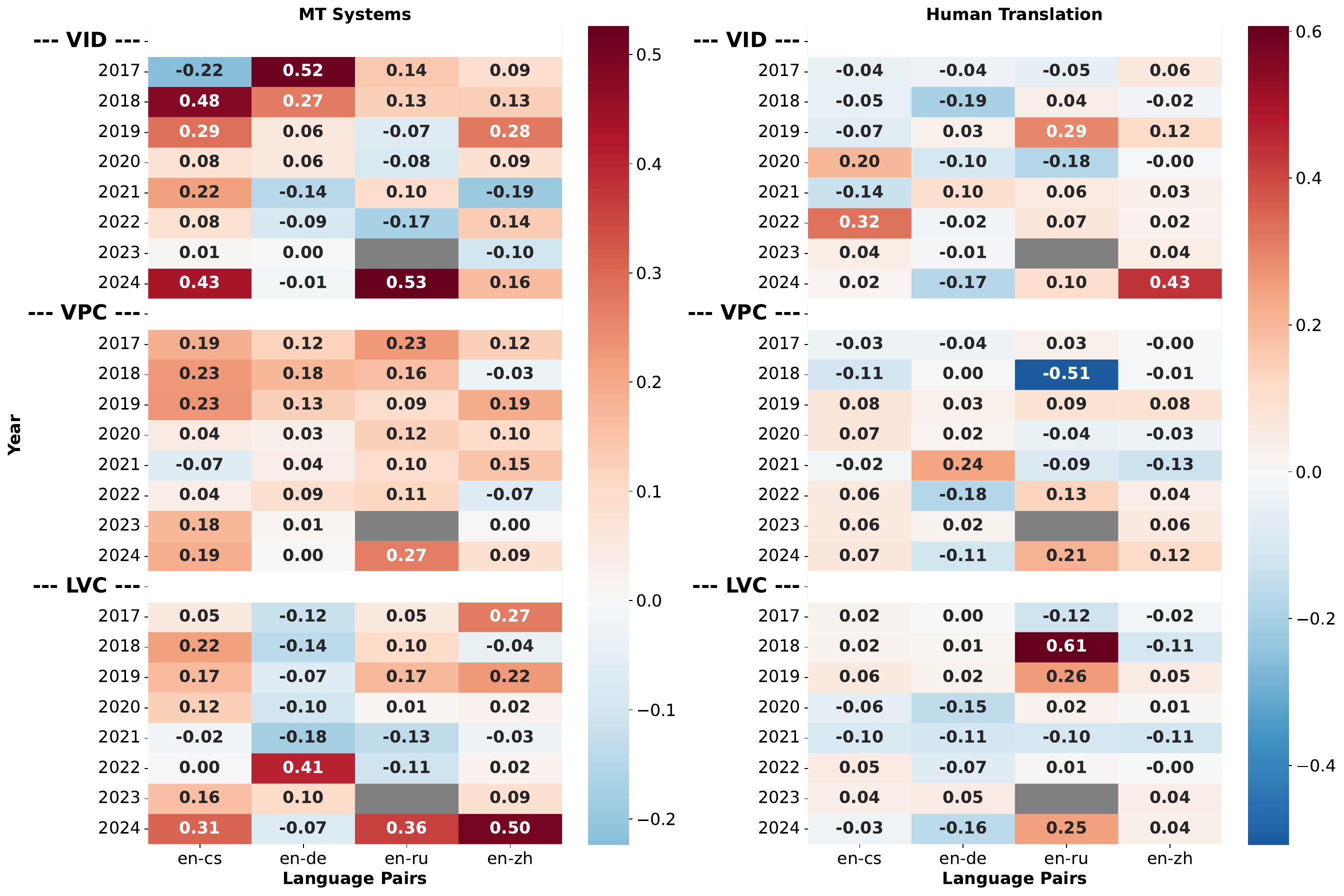}
    \caption{Comparison of the translation quality between sentences with and without VMWE, using the human DA scores (\textit{typically range from -1 to 1, higher the better}). Left is machine translation and right is human translation (both evaluated by humans). For each cell, the values are calculated as the score of non-VMWE sentences minus VMWE sentences. Grey cell indicates missing language pair in that year. Anomalies for Human DA scores on en--ru 2018 are explained in appendix~\ref{Anomalies}.}
    \label{fig:heatmap_metric_test}
\end{figure*}

\subsection{Performance on VMWE Datasets}

Figure~\ref{fig:heatmap_metric} provides a comprehensive comparison of VMWE effects on MT system performance using existing VMWE datasets, evaluated by MetricX and xCOMET as discussed in Section \ref{existing_datasets}. We evaluate the QE scores of sentences with and without the language phenomenon (VMWE vs. non-VMWE), and report the difference $\delta$ between them. For MetricX-24 (lower the better), $\delta_{\text{MetricX24}}=MetricX24_{\text{VMWE}} - MetricX24_{\text{non-VMWE}}$.

Since xCOMET is higher the better, we reverse it to make it consistent: $\delta_{\text{xCOMET}}=xCOMET_{\text{non-VMWE}} - xCOMET_{\text{VMWE}}$. Positive values indicate VMWE sentences have a worse translation quality than non-VMWE sentences. We observe that VMWEs consistently degrade translation quality across multiple MT systems and languages. Verbal idioms experience the largest performance gap, followed by verb-particles and light verb constructions.

A language effect is also visible: In the MetricX view, penalties tend to be larger for en--cs and en--ru than for en--zh, with en--de in between—especially pronounced for VID. xCOMET has a different view: en--cs and en--zh tend to bear larger penalties than en--de. Overall, the two metrics agree in direction and relative magnitude: VID shows the largest degradation, moderately on VPC, then LVC. The performance of remaining three language pairs is consistent with this trend, details in Appendix~\ref{app:Rest 3 delta}.

\noindent \textbf{Invalid Translations}. Some MT models, particularly the LLM-based ones, frequently produce invalid outputs---wrong target language, empty translation, or repetitive text. We exclude these instances from our reported scores. GemmaX2 mistranslates over 75\% of \textit{en-cs} sentences, so we omit its results for that pair (grey cell). We further discuss these mistranslations in Appendix~\ref{invalid_translations}, and provide error cases in Appendix~\ref{error_analysis}. 

\paragraph{Takeaway.} VMWEs consistently degrade translation quality across all tested models. This penalty scales directly with non-compositionality, hitting verbal idioms the hardest.



\subsection{Performance on WMT Dataset}

\noindent \textbf{MT Systems' Performance on Sentences w/ and w/o VMWEs.} After extracting the VMWE sentences from the WMT dataset, we compare their human DA scores with those of non-VMWE sentences of each year's top four MT systems from 2017 to 2024. Positive scores mean that VMWE sentences received lower score than non-VMWE. In the left heatmap of Figure~\ref{fig:heatmap_metric_test}, top MT systems are still negatively affected by the presence of VMWE. Notably, this trend aligns with our earlier findings using MetricX-24 and xCOMET, though the previous results are more salient and consistent. It suggests that both automatic evaluation and human judgment confirm the translation degradation.

\noindent \textbf{Human Translators' Performance on Sentences w/ and w/o VMWEs.} The WMT datasets also provide multiple human translators' translations (also evaluated by human judges). From the right heatmap in Figure~\ref{fig:heatmap_metric_test}, English to Russian is most affected overall. In contrast to machine translation, human translations appear to be less affected by VMWEs. Most cells show a small delta, indicating that humans translate VMWE sentences with comparable quality to non-VMWE ones. This human--machine contrast is consistent with our earlier metric-based findings. 


\paragraph{Takeaway.} Unlike humans, current MT systems consistently fail to fully capture the underlying figurative semantics in VMWEs, exposing a critical gap in contextual understanding.



\section{Analysis}

Our results show VMWE sentences demonstrate degradation in overall translation quality across languages and MT systems. However, an important question remains: are there any confounding factors that contribute to this degradation? It may be possible that the VMWE sentences might just be harder sentences in general, so the observed degradation may not be due to multiword expression. In this section, we perform deeper analysis and show the presence of the VMWE candidate is the primary cause for the translation quality degradation.

\subsection{xCOMET Error Span}
xCOMET provides token-level error spans in the translated text marked with severity labels. For this experiment, we first find the overlap percentage between these error spans and the VMWE candidate in the translated text. We then show that this overlap percentage has a high correlation with the QE score assigned by xCOMET. This will help identify if the degradation in translation quality is primarily due to the presence of the VMWE candidate. Since xCOMET outputs the error span in the translated text, a key part of this analysis is to align the tokens between the English source sentence and its translation. For this, we use \textsc{simalign}~\citep{jalili-sabet-etal-2020-simalign} to obtain bilingual word alignments. \textsc{simalign} computes contextual embeddings for source and target tokens with a multilingual Transformer, constructing a similarity matrix via cosine similarity between every source–target token pair, then extracting alignments using heuristic matching. We then use these alignments to trace errors back to the source-side VMWE candidate.

\begin{table}[t]
    \centering
    \scriptsize
    \resizebox{0.99\linewidth}{!}{
    \begin{tabular}{l|c|c|c|c}
    \toprule
    \textbf{MT System} & \textbf{VID} & \textbf{VPC} & \textbf{LVC} & \textbf{xCOMET}\\
    \midrule
     Opus & 78.64 & 65.51 & 62.21 & 66.89 \\
     M2M & 82.15 & 67.67 & 62.19 & 73.02 \\
     Phi-4-multi & 66.50 & 54.43 & 56.36 & 74.68 \\
     Madlad & 69.88 & 52.79 & 50.14 & 78.22 \\
     Seamless & 67.91 & 56.30 & 55.75 & 76.43 \\
     LLaMAX & 66.82 & 53.11 & 55.89 & 78.08 \\
     GemmaX2 & 55.41 & 46.77 & 43.90 & 85.69 \\
     Google API & \textbf{52.98} & \textbf{40.77} & \textbf{39.25} & \textbf{87.19}\\
     \bottomrule
    \end{tabular}
    }
\caption{Percentage of sentences in which xCOMET-QE error spans overlap the source VMWE span. Values are averaged over seven target languages. Lower overlap indicates errors are less concentrated on the verb multiword expression itself. Stronger systems show relatively lower overlap, while weaker systems exhibit higher overlap, especially on verbal idioms.}
\label{table:overlap_percentage}
\end{table}


Table~\ref{table:overlap_percentage} shows error span overlap for eight models across seven language pairs. A high error span overlap indicates the MT system struggles with the verbal multiword expressions. The two best performing models, GemmaX2 and Google API consistently have low overlap percentage, while the MT systems that degrade the most like Madlad or Opus show a high score, especially for VIDs. See Appendix~\ref{ESO} for details about error span overlap.

\paragraph{Takeaway.} Token-level alignments confirm errors localize directly on VMWE spans, with severity strongly correlating with the expression's degree of non-compositionality.



\subsection{Confounding Factors}
A potential concern in comparing VMWE and non-VMWE segments is that the observed quality differences may reflect general sentence difficulty rather than effects specific to VMWE phenomena. VMWE-containing sentences may be longer, more lexically ambiguous, or syntactically more complex, all of which are known to negatively impact machine translation quality. To account for this confound, we perform a segment-level regression analysis that provides controls for standard source-side difficulty factors.

Following \citet{araghi2024}, we compute four features for each English source sentence: (i) if the sentence contains VMWE ($I_{vmwe}$), (ii) sentence length ($S_{len}$): number of tokens, (iii) degree of polysemy ($P_{deg}$): average number of WordNet senses per content word \citep{miller-1992-wordnet}, and (iv) structural complexity ($T_{cmp}$): sum of dependency arc lengths parsed by spaCy \citep{spacy2020}. All continuous features are z-normalized. We then fit linear regression models with QE scores as dependent variables and VMWE presence as the main predictor, alongside the three difficulty features. Specifically, we estimate segment-level linear models of the form:
\[
\text{score}_i = \beta_0 + \mathbf{x}_i^\top \boldsymbol{\beta} + \varepsilon_i,
\]
where $\mathbf{x}_i = [I_{vmwe,i},\, S_{len,i},\, P_{deg,i},\, T_{cmp,i}]$, and $\varepsilon_i$ denotes the residual error term.

The regression dataset consists of $N = 305{,}428$ segment-level observations drawn from our QE experiments. Each observation corresponds to a single translation output produced by a particular MT system, for a given English source sentence, into a specific target language. We pool across all MT systems and all language pairs used in our experiments, and retain only segments for which both QE scores (xCOMET and MetricX) and all three difficulty features are available. This yields 205,565 VMWE outputs and 99,863 non-VMWE outputs from the BNC baseline. 

\begin{table}[t]
    \centering
    \resizebox{0.95\linewidth}{!}{
    \begin{tabular}{l|c|l}
    \toprule
    \textbf{Predictor} & \textbf{xCOMET $\beta$ (SE)} & \textbf{MetricX $\beta$ (SE)} \\
    \midrule
    $\beta_0$ & 0.7908 (0.0010)$^{***}$ & 5.6889 (0.0183)$^{***}$ \\
    $I_{vmwe}$ & -0.0813 (0.0012)$^{***}$ & 0.9954 (0.0240)$^{***}$ \\
    $S_{len}$ & -0.0359 (0.0012)$^{***}$ & 0.6348 (0.0240)$^{***}$ \\
    $P_{deg}$ & -0.0126 (0.0006)$^{***}$ & 0.0244 (0.0110)$^{*}$ \\
    $T_{cmp}$ & -0.0120 (0.0009)$^{***}$ & 0.0059 (0.0210) \\
    \bottomrule
    \end{tabular}
    }
\caption{Segment-level linear regression of QE scores on VMWE presence and source-side difficulty features. Standard errors are in parentheses. $\beta_0$ is the expected score for non-VMWE segments at mean difficulty. For MetricX, higher $\text{score}_i$ indicates worse quality.
Significance codes: $^{***}p<0.001$, $^{**}p<0.01$, $^{*}p<0.05$.}
\label{table:confounds_regression}
\end{table}


Table~\ref{table:confounds_regression} reports the estimated coefficients $\beta$ with standard errors in parentheses; for the z-scored difficulty predictors, $\beta$ corresponds to a one-standard-deviation increase in the feature. Model fit statistics indicate modest but expected explanatory power at the sentence level, with $R^2 = \text{adjusted } R^2 = 0.063$ for xCOMET and $R^2 = \text{adjusted } R^2 = 0.027$ for MetricX. As expected, longer, more polysemous and structurally complex sentences are harder for MT, exhibiting lower xCOMET scores and higher MetricX scores. Importantly, VMWE presence remains a strong predictor of translation degradation after controlling for these difficulty factors. On average, VMWE sentences receive xCOMET scores approximately 0.08 points lower on the 0--1 scale, and MetricX scores about 1 point higher (worse) on the 0--25 scale. Taken together, VMWE presence remains a significant predictor of QE scores conditional on other factors.

\paragraph{Takeaway.} Controlling for general sentence difficulty confirms the translation degradation stems directly from VMWE presence, rather than broader linguistic complexity.



\begin{table}[t]
    \centering
    \resizebox{0.95\linewidth}{!}{
    \begin{tabular}{ll|ccc}
    \toprule
    \textbf{Type} & \textbf{Lang} & \textbf{GPT-4.1} & \textbf{GPT-5.1} & \textbf{Google API} \\
    \midrule
\multirow{5}{*}{\textbf{VID}} & en--cs & $+0.18$ & $+0.22$ & $+0.25$ \\
& en--de & $+0.01$ & $+0.02$ & $+0.05$ \\
& en--ru & $+0.05$ & $+0.06$ & $+0.08$ \\
& en--zh & $+0.16$ & $+0.14$ & $+0.12$ \\
& \textit{Avg} & $+0.10$ & $+0.11$ & $+0.13$ \\
\midrule
\multirow{5}{*}{\textbf{VPC}} & en--cs & $+0.01$ & $+0.00$ & $+0.12$ \\
& en--de & $-0.01$ & $+0.00$ & $+0.02$ \\
& en--ru & $+0.01$ & $+0.00$ & $+0.03$ \\
& en--zh & $+0.04$ & $+0.02$ & $+0.03$ \\
& \textit{Avg} & $+0.01$ & $+0.00$ & $+0.05$ \\
\midrule
\multirow{5}{*}{\textbf{LVC}} & en--cs & $+0.00$ & $-0.01$ & $+0.04$ \\
& en--de & $-0.02$ & $+0.00$ & $+0.00$ \\
& en--ru & $+0.01$ & $+0.01$ & $+0.01$ \\
& en--zh & $+0.04$ & $+0.02$ & $+0.01$ \\
& \textit{Avg} & $+0.01$ & $+0.00$ & $+0.02$ \\
    \bottomrule
    \end{tabular}
    }
\caption{On the 100-sentence-per-category sampled subset. We report $\delta_{\text{xCOMET}} = xCOMET_{\text{non-VMWE}} - xCOMET_{\text{VMWE}}$. Larger values indicate worse performance on VMWE sentences than non-VMWE ones.}

\label{tab:xcomet_gpt41_gpt51}
\end{table}

\begin{table*}[t]
\small
\centering
\setlength{\tabcolsep}{4pt}
\renewcommand{\arraystretch}{1.08}
\begin{tabular}{@{}p{0.07\textwidth}p{0.22\textwidth}p{0.25\textwidth}p{0.36\textwidth}@{}}
\toprule
\textbf{Cat.} & \textbf{Source} & \textbf{Translation} & \textbf{Explanation} \\
\midrule

\textbf{LVC} &
``Time we got some sleep,'' said Thomas. &
\foreignlanguage{russian}{«Время, когда мы \err{немного заснули}», — сказал Томас.} &
The error is \foreignlanguage{russian}{заснули} (“fell asleep”), which translates “got some sleep” too literally and misses the intended meaning that it is time for them to go to sleep.\\
\midrule

\textbf{VPC} &
Breaking up, as Neil Sedaka once put it, is hard to do. &
\textit{\err{Romper}, como dijo una vez Neil Sedaka, es difícil de hacer.} &
The error is the word \textit{Romper} (“to break”), which translates \textit{breaking up} literally and misses the sense of ending a romantic relationship. \\
\midrule

\textbf{VID} &
``He's dead to the world,'' said Mabel, over her shoulder. &
\begin{CJK*}{UTF8}{gbsn}他已经\textcolor{red}{死了}，``梅布尔回头说道。''\end{CJK*}&
The error is \begin{CJK*}{UTF8}{gbsn}死了\end{CJK*} (“dead”), which renders \textit{dead to the world} literally and loses its idiomatic meaning “fast asleep.” \\
\bottomrule
\end{tabular}

\caption{Examples of VMWE translation errors for the three subtypes in this study.}
\label{tab:error_analysis}
\end{table*}

\subsection{Comparison with Top-tier LLMs}
\label{sec:top_tier_llm}

In addition to the eight primary MT systems evaluated in our main experiments, we further evaluate top-tier LLMs as direct translators on a sampled subset to test whether the VMWE effect persists even for stronger general-purpose models. Prior work and recent multilingual MT results suggest that modern LLM-based translators and commercial MT APIs can be competitive with specialized MT systems on general translation benchmarks~\citep{cui2025multilingualmachinetranslationopen}. Verifying performance specifically on VMWE-containing inputs therefore provides a stronger test of our main claim.


For this additional experiment, we run GPT-4.1~\citep{openai2024gpt4technicalreport}, GPT-5.1~\citep{singh2025openaigpt5card} and Google Translate API on the same sampled subset: 100 sentences per category (\textsc{non-VMWE}, \textsc{VID}, \textsc{VPC}, \textsc{LVC}), translating English into Czech, German, Russian, and Chinese. The translation prompt for GPT-4.1 and GPT-5.1 is in Appendix~\ref{app:gpt_translation_prompt}. We score the outputs with the same xCOMET evaluation pipeline and define $\delta_{\text{xCOMET}} = \text{xCOMET}_{\text{non-VMWE}} - \text{xCOMET}_{\text{VMWE}}$. Larger values indicate worse performance on VMWE sentences. MetricX-24 results on the same sample are reported in Appendix~\ref{app:top_tier_llm_metricx}.

Table~\ref{tab:xcomet_gpt41_gpt51} shows that the VMWE effect persists even for top-tier LLMs. First, \textsc{VID} remains consistently more difficult than \textsc{non-VMWE} across all three systems and all four language pairs, indicating that the core difficulty of highly non-compositional verbal idioms is not eliminated by stronger LLM-based translation. Second, the degradation for \textsc{VPC} and \textsc{LVC} is much smaller, and in some cases near zero, especially for GPT-4.1 and GPT-5.1. This mirrors our broader finding that VMWE difficulty follows a compositionality gradient, with verbal idioms showing the largest degradation. Google Translate API generally exhibits larger positive gaps than GPT-4.1 and GPT-5.1, particularly for \textsc{VID} and \textsc{VPC}.

\paragraph{Takeaway.} Though top-tier LLMs reduce translation degradation for semi-compositional expressions, they still struggle significantly with highly non-compositional verbal idioms.



\subsection{Error Analysis}

Table~\ref{tab:error_analysis} gives one example for each VMWE subtype: LVC, VPC, and VID.


In the LVC example, the problem comes from translating \textit{got some sleep} too literally. In this sentence, the phrase means “sleep for a while” or “get some rest,” but the Russian version gives a meaning closer to “fell asleep a little,” which sounds awkward to a native speaker.


The VPC example has a similar issue. The system translates \textit{breaking up} as \textit{romper}, but that does not clearly express the idea of ending a romantic relationship. The result is understandable, but it does not sound like a natural translation.

The VID example is more serious. The expression \textit{dead to the world} is an idiom meaning “deeply asleep,” but the Chinese translation takes it literally and says that the person is dead. This changes the meaning of the sentence completely.

\paragraph{Takeaway.} VMWE-specific errors primarily stem from literal processing of non-compositional phrases, yielding outputs that either sound unnatural or completely alter the intended meaning.



\section{Conclusion}
Verbal multiword expressions systematically degrade machine translation quality. Across languages and systems, we observe a consistent gradient: verbal idioms are most disruptive, verb--particle constructions show moderate degradation, and light-verb constructions are least affected. This trend holds under both MetricX-24 and xCOMET. The degradation is primarily attributable to VMWE presence rather than general sentence-level difficulty. Together, these findings suggest that improving robustness on everyday text requires explicit handling of non-compositionality---through targeted VMWE detection, VMWE-aware training, and selective pre-editing or normalization---to produce translations that are more faithful and reliable in the presence of multiword expressions.




\section*{Limitations}

While we conduct a large-scale evaluation of VMWEs, our study comes with some limitations. The WMT datasets have human reference scores for the translations, but the translated sentences in VMWE datasets are solely judged by QE models and do not involve human reference scores. While the QE translation scores align with the human reference scores according to our experiments, it still remains a limitation in our work.

Our work involves a large-scale analysis of the negative impact of verbal multiword expressions (VMWEs) on machine translation quality. While our findings highlight important details about how different VMWE categories and languages are affected, we do not propose any new method to improve translation quality.

Our evaluation focuses on verbal multiword expressions. However, other non-literal language phenomena, such as metaphor and metonymy, pose similar challenges of non-literal interpretation for NLP systems~\citep{ghosh-jiang-2025-conmec, ghosh2026metfuse} and likely affect MT quality in related ways. Extending our framework to cover these phenomena is a promising direction for future work.

Finally, our scope is limited from English to Czech, German, Spanish, Japanese, Russian, Turkish, and Chinese. We do not explore low-resource languages and other translation directions.

\section*{Acknowledgments}

This work is supported in part by a URC Faculty Scholars Research Award from the Office of Research at the University of Cincinnati. We thank the CincyNLP group for their feedback. We also thank the anonymous ACL reviewers for their insightful suggestions.

\bibliography{colm2025_conference}

\newpage
\appendix

\section{Machine Translation Systems Used}
\label{mt_systems}

We used the following widely used machine translation systems in this work: \\
\textbf{1) Madlad400}~\citep{kudugunta2023madlad400}: A Google multilingual machine translation model, based on the T5 architecture~\citep{t5_paper} that was trained on 250 billion tokens covering over 450 languages. The model we used here is madlad400-10b-mt.\\
\textbf{2) SeamlessM4T}~\citep{seamless2023}: Meta AI's massively multilingual and multimodal machine translation model, supporting an impressive range of translation capabilities with 96 languages for text input/output. The model we used here is seamless-m4t-v2-large, model size 2.3B.\\
\textbf{3) M2M100}~\citep{fan2020englishcentric}: A Meta-powered multilingual encoder-decoder model, primarily designed for translation tasks, supporting direct translation between 100 languages without requiring English as an intermediate language. The model we used here is m2m100-1.2B.\\
\textbf{4) Opus-MT}~\citep{tiedemann-thottingal-2020-opus}: Provides open translation services built on the Marian neural machine translation framework~\citep{marian_translation}, trained on Opus data, and later converted to PyTorch models for the Hugging Face ecosystem. The models we used here are from Helsinki-NLP---opus-mt-en-xx, "xx" is the target language code, mdoel size are all less than 1B.\\
\textbf{5) LLaMAX3 Alpaca}~\citep{lu-etal-2024-llamax}: An LLM-based machine translation model, LLaMAX combines powerful multilingual translation capabilities with instruction-following abilities. This model extends Meta's LLaMA 3 architecture~\citep{llama3} to support translation between over 100 languages without sacrificing its ability to follow complex instructions. The model we used here is LLaMAX3-8B-Alpaca.\\
\textbf{6) Phi-4-multimodal}~\citep{microsoft2025phi4minitechnicalreportcompact}: Built upon the pretrained Phi-4-mini model, Phi-4-multi can processes text, image, and audio inputs to generate text outputs. While primarily known for its multimodal capabilities, it can handle translation tasks as part of its broader language understanding abilities, making it another LLM-based MT system for our task. The model we used here is Phi-4-multimodal-instruct, model size 5.6B.\\
\textbf{7) GemmaX2}~\citep{cui2025multilingualmachinetranslationopen}: A very recent multilingual LLM-based translation model, that achieved state-of-the-art performance across 28 languages. Based on Google's Gemma2 architecture~\citep{gemma2}, it consistently outperforms other LLM-based MT models like TowerInstruct and XALMA, achieving competitive results with Google Translate and GPT-4-turbo. The model we used here is GemmaX2-28-9B-v0.1.\\
\textbf{8) Google Translate API}\footnote{\url{https://cloud.google.com/translate?hl=en}}: A widely used multilingual neural machine translation service developed by Google, offering translations for 249 languages and language varieties as of March 2025. \\

We run all MT models on a server with eight NVIDIA RTX 6000 Ada Generation GPUs.

\section{Invalid Translations by LLM-Based MT Models}
\label{invalid_translations}

During evaluation, we observed several instances of invalid outputs from LLM-based MT models. The most common cases were translations in the wrong language (e.g., translating to Russian text when asked for Chinese), and untranslated sentences where the model outputs the English source sentence. We also observed some cases of malformed or repetitive output, such as repeated characters or phrases in the translated text. To identify invalid translations, we use the lingua-language-detector-2.1.0\footnote{\url{https://pypi.org/project/lingua-language-detector/}} to automatically detect the language of the translated output and match it with the intended target language. If there is a mismatch, the translation is marked as invalid. We filter out these invalid translations when reporting the scores for VMWE vs non-VMWE sentences. 

\begin{table}
\centering
\resizebox{0.95\linewidth}{!}{
    \begin{tabular}{l|c|c}
    \toprule
    MT System & Language Pair & Percentage of Errors \\
    \midrule
    LlaMAX & en-tr & 99.89 \\
    LlaMAX & en-ja & 99.80 \\
    LlaMAX & en-de & 16.03 \\
    LlaMAX & en-cs & 14.01 \\
    \midrule
    GemmaX2 & en-ja & 78.68 \\
    GemmaX2 & en-cs & 75.69 \\
    GemmaX2 & en-tr & 74.91 \\
    GemmaX2 & en-es & 53.16 \\
    GemmaX2 & en-de & 14.57 \\
    GemmaX2 & en-zh & 14.06 \\
    GemmaX2 & en-ru & 11.44 \\
    \midrule
    Opus & en-tr & 62.74 \\
    \bottomrule
    \end{tabular}
    }
\caption{Percentage of mistranslated sentences.}
\label{table:error_percentage}
\end{table}

\begin{table*}
    \centering
    \resizebox{0.6\linewidth}{!}{
    \begin{tabular}{lcccccc}
    \toprule
    \multirow{2}{*}{\textbf{Models}} & \multicolumn{2}{c}{\bf LVC} & \multicolumn{2}{c}{\bf VPC} & \multicolumn{2}{c}{\bf VID} \\
    \cmidrule(lr){2-3} \cmidrule(lr){4-5} \cmidrule(lr){6-7}
    & Pos & Neg & Pos & Neg & Pos & Neg \\
    \midrule
    GPT-4o & \textbf{81.6} & \textbf{78.0} & 80.0 & 84.4 & 81.8 & 78.9 \\
    Phi-4  & 71.3 & 70.6 & \textbf{80.5} & \textbf{84.9} & 81.7 & \textbf{80.1} \\
    Llama-3.3-70B & 56.7 & 44.7 & \textbf{80.5} & \textbf{84.9} & \textbf{83.1} & 79.4 \\
    DeepSeek-R1-70B & 55.8 & 57.1 & 77.6 & 83.3 & 80.5 & 79.3 \\
    \bottomrule
    \end{tabular}
    }
\caption{F1-score comparison of different LLMs. Pos: VMWE sentences. Neg: non-VMWE sentences.}
\label{table:llm_comparison}
\end{table*}

Table \ref{table:error_percentage} shows all MT systems and corresponding languages pairs where the percentage of mistranslated sentence in all existing VMWE datasets is greater than 10. LlaMAX fails to convert more than 99\% of the sentences to the Turkish and Japanese. While GemmaX2 achieves SOTA performance in terms of QE score, it suffers consistently from mistranslating---all 7 languages have significant error rates, with Japanese, Czech and Turkish have over 70\% errors. Interestingly, Opus is the only MT system with a high error rate that is not LLM-based (we use model Helsinki-NLP/opus-mt-en-trk to translate English to Turkish). 

\section{MetricX-24 Results for Top-tier LLMs on the Controlled Sample}
\label{app:top_tier_llm_metricx}

For completeness, we also report MetricX-24 degradation gaps on the same sampled subset of 100 sentences per category. Since lower MetricX-24 indicates better translation quality, we define $\delta_{\text{MetricX24}} = MetricX24_{\text{VMWE}} - MetricX24_{\text{non-VMWE}}$, where larger values indicate worse performance on VMWE sentences.

\begin{table}[t]
    \centering
    \resizebox{0.90\linewidth}{!}{
    \begin{tabular}{ll|ccc}
    \toprule
    \textbf{Type} & \textbf{Lang} & \textbf{GPT-4.1} & \textbf{GPT-5.1} & \textbf{Google API} \\
    \midrule
\multirow{5}{*}{\textbf{VID}} & en--cs & $+0.95$ & $+0.48$ & $+2.00$ \\
& en--de & $+0.68$ & $+0.34$ & $+0.45$ \\
& en--ru & $+0.70$ & $+0.42$ & $+0.70$ \\
& en--zh & $+0.62$ & $+0.45$ & $+0.52$ \\
& \textit{Avg} & $+0.74$ & $+0.42$ & $+0.92$ \\
\midrule
\multirow{5}{*}{\textbf{VPC}} & en--cs & $+0.61$ & $-0.05$ & $+1.48$ \\
& en--de & $+0.43$ & $+0.20$ & $+0.38$ \\
& en--ru & $+0.50$ & $-0.01$ & $+0.50$ \\
& en--zh & $+0.63$ & $+0.23$ & $+0.38$ \\
& \textit{Avg} & $+0.54$ & $+0.09$ & $+0.68$ \\
\midrule
\multirow{5}{*}{\textbf{LVC}} & en--cs & $+0.10$ & $-0.86$ & $+0.40$ \\
& en--de & $+0.06$ & $-0.14$ & $+0.06$ \\
& en--ru & $+0.38$ & $+0.15$ & $+0.16$ \\
& en--zh & $+0.68$ & $+0.24$ & $+0.33$ \\
& \textit{Avg} & $+0.31$ & $-0.15$ & $+0.24$ \\
    \bottomrule
    \end{tabular}
    }
\caption{On the 100-sentence-per-category sampled subset. We report $\delta_{\text{MetricX24}} = MetricX24_{\text{VMWE}} - MetricX24_{\text{non-VMWE}}$. Larger values indicate worse performance on VMWE sentences than non-VMWE ones.}

\label{tab:gpt41_gpt51_google}
\end{table}

Table~\ref{tab:gpt41_gpt51_google} shows a pattern broadly consistent with the xCOMET results in the main text. \textsc{VID} remains the most consistently difficult category across systems and languages. GPT-5.1 generally reduces the average degradation gap relative to GPT-4.1, especially for \textsc{VPC} and \textsc{LVC}, although some values become small or negative on this 100-sentence-per-category sample. We interpret these near-zero or negative gaps cautiously, as they may partly reflect sampling variance in a relatively small controlled subset. Even so, the overall pattern remains consistent with our main findings: stronger LLMs reduce the VMWE degradation gap, but do not eliminate it, particularly for verbal idioms.

\section{Anomalies in WMT Human Translation Score}
\label{Anomalies}
On closely examining Figure~\ref{fig:heatmap_metric_test}, we identify English–Russian (en–ru) in 2018 as an outlier in the human translation scores.

We investigated these specific anomalies and found that they are largely due to data sparsity and uneven sampling in the WMT human evaluation logs for that year. In the 2018 en–ru dataset, only a very small number of reference translations received human DA scores: 243 evaluated reference segments in total, of which just 11 contain LVCs and 10 contain VPCs. By contrast, the 2019 en–ru dataset has 1,499 evaluated reference segments, including 37 LVC sentences and 105 VPC sentences.

\section{Qualitative Discussion}

A consistent finding across our experiments is that \textsc{VID} is the most affected VMWE category in machine translation, followed by \textsc{VPC}, while \textsc{LVC} is affected the least. This ordering is stable across our main experimental settings. In a supplementary check on a small controlled subset (100 sentences per category), GPT-4.1 and GPT-5.1 exhibit the same qualitative pattern under both MetricX-24 and xCOMET, which is consistent with the broader trends observed for the primary baselines, but we do not treat this small-scale LLM comparison as a standalone conclusion.

We interpret the overall pattern through the lens of compositionality. \textsc{LVC}s are often more compositional: much of the semantic content is carried by the noun while the verb is semantically light (e.g., ``\textit{take a walk}''). As a result, translation is less sensitive to the particular light verb choice and errors tend to resemble minor lexical substitutions rather than meaning shifts. \textsc{VPC}s are typically semi-compositional, where the particle may be literal in some contexts but idiomatic in others, leading to intermediate difficulty. \textsc{VID}s are strongly non-compositional, and their meanings cannot be recovered from their parts (e.g., ``\textit{beat the clock}'' meaning ``finish before the deadline''). Correct translation therefore requires selecting a target-language expression that conveys the intended meaning rather than a literal rendering, which remains challenging across systems and languages. Overall, our results support the conclusion that \textbf{VMWE-related degradation in machine translation correlates with the degree of non-compositionality}.

\section{Heatmap for Additional Languages}
\label{app:Rest 3 delta}

\begin{figure*}[t]
    \centering
    \includegraphics[width=0.9\linewidth]{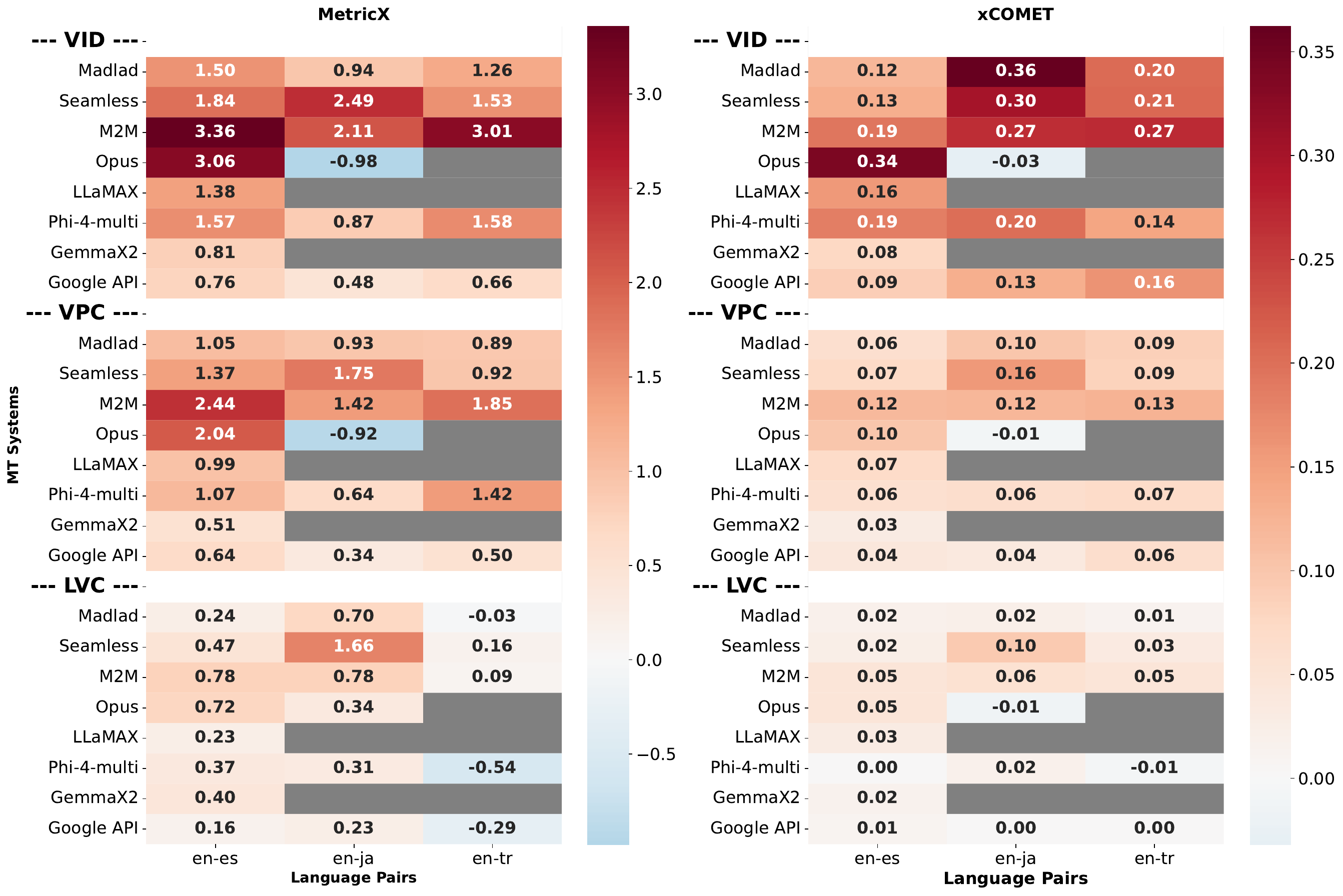}
    \caption{Comparison of the translation quality between sentences with and without VMWE, using MetricX-24-QE (\textit{range from 0 to 25, lower the better}) and xCOMET-QE (\textit{range from 0 to 1, higher the better}) scores. For MetricX-24 (left), the cell values are calculated as the difference of averaged score (VMWE minus non-VMWE). For xCOMET (right), the scores are opposite as non-VMWE minus VMWE. Larger numbers (darker colors) mean worse translation performance on VMWE sentences.}
    \label{fig: delta 3}
\end{figure*}

Reading Figure~\ref{fig: delta 3}, sentences containing VMWEs are consistently penalized across systems and language pairs, with a clear gradient on both Metricx-24 and xCOMET.

\noindent\ Language effects. Across both metrics, en--es tends to show the largest gaps, en--tr is moderate, and en--ja often attenuates or flips some systems, especially for VID and VPC. Overall, the two metrics agree in direction and relative magnitude, reinforcing the non-compositionality gradient.

\section{Human Direct Assessment (DA)}
\label{app:da}

Direct Assessment (DA) is WMT’s segment-level human evaluation protocol. Raters view a source sentence, a single system translation and human-provided reference translation. Then assign a scalar quality score on a continuous scale in 0–100, reflecting overall adequacy/fluency.

\paragraph{How WMT collects DA.}
Each segment receives multiple independent ratings. Items are shown one-at-a-time; raters use a slider to score quality. WMT includes control items and standard filters to remove unreliable ratings.

\paragraph{How DA scores are handled.}
Raw scores are standardized \emph{per rater} to remove scale bias:
\[
z_{i,j} = \frac{s_{i,j} - \mu_j}{\sigma_j},
\]
where $s_{i,j}$ is rater $j$’s score for segment $i$, and $\mu_j,\sigma_j$ are that rater’s mean and standard deviation. We obtain subset or system scores by averaging $z_{i,j}$ over relevant segments (and raters), optionally with stratified bootstrap for uncertainty.

\paragraph{How we use DA in this paper.}
We automatically identify English source sentences containing a VMWE (VID/VPC/LVC) and construct a non-VMWE comparison subset. For each language pair, year, and system, we collect corresponding DA separately for VMWE and non-VMWE sentences and report their difference to assess VMWE impact under human judgment.

\section{Comparing LLM performance in Extracting VMWE Sentences}
\label{app:llm_comparison_sec}

The Conference on Machine Translation (WMT) dataset contains English sentences along with their corresponding translations in other languages by MT systems and humans, accompanied by human judgment (direct assessment scores)~\citep{graham-etal-2013-continuous} evaluating the translation quality. Our goal in section \ref{section_3} is use this dataset to evaluate if human translation scores align with the MT evaluation models we used in section \ref{section_2}. The VMWE candidate sentences are collected by these rules: For verbal idioms, we use BLEU-4~\citep{papineni-etal-2002-bleu}, threshold of 0.6 to find the verbal idiom candidates based on the verbal idiom dictionary from EPIE and MAGPIE. For verb-particle constructions, we use spaCy dependency parser~\citep{spacy2020} where a particle has a \textit{prt} relation to a verb to extract the verb-particle construction candidates. And light verb constructions are detected using the light verb lists (\textit{have, take, make, get, put, give, pay, do, offer, raise}), the related noun are extracted by spaCy dependency parser. After that, we use GPT-4o to extract VMWE sentences.

For the extraction, we tested different LLMs on the existing VMWE datasets. Table \ref{table:llm_comparison} shows the performance of different LLMs in classifying VMWE on existing datasets. We test 4 different models, GPT-4o~\citep{openai2024gpt4ocard}, Phi-4~\citep{microsoft2025phi4minitechnicalreportcompact}, Llama 3.3-70B~\citep{llama3} and DeepSeek-R1-Distill-70B~\citep{deepseek}. We use the same prompts and few-shot examples for all models, given in Appendix \ref{app:llm-extract}. All the LLMs have nearly identical performance in VPC and VID categories, with Phi-4 and Llama having exact same F1 scores in VPC. However, GPT-4o establishes dominance in LVC category, with all other LLMs performing significantly worse. Due to its balanced performance across all categories, we chose GPT-4o for the main experiment.

\section{Comparing Translation Quality across Languages \& MT Systems}

\begin{figure}[t]
    \centering
    \includegraphics[width=0.95\linewidth]{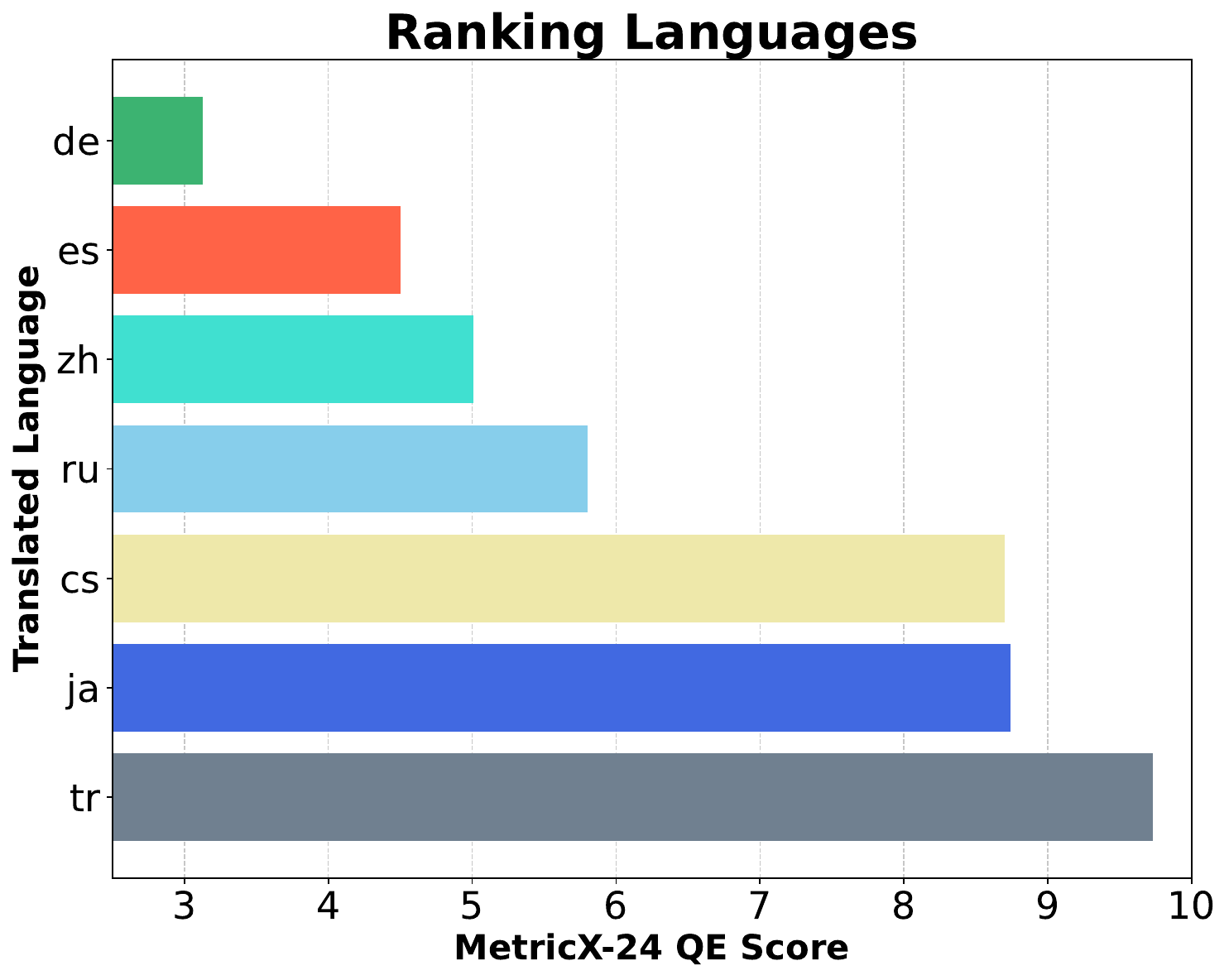}
    \caption{Ranking languages based on translation quality for sentences with VMWE using MetricX-24. Lower score indicates better performance.}
    \label{fig:language_rank_diagram}
\end{figure}

\begin{figure}[t]
    \centering
    \includegraphics[width=0.95\linewidth]{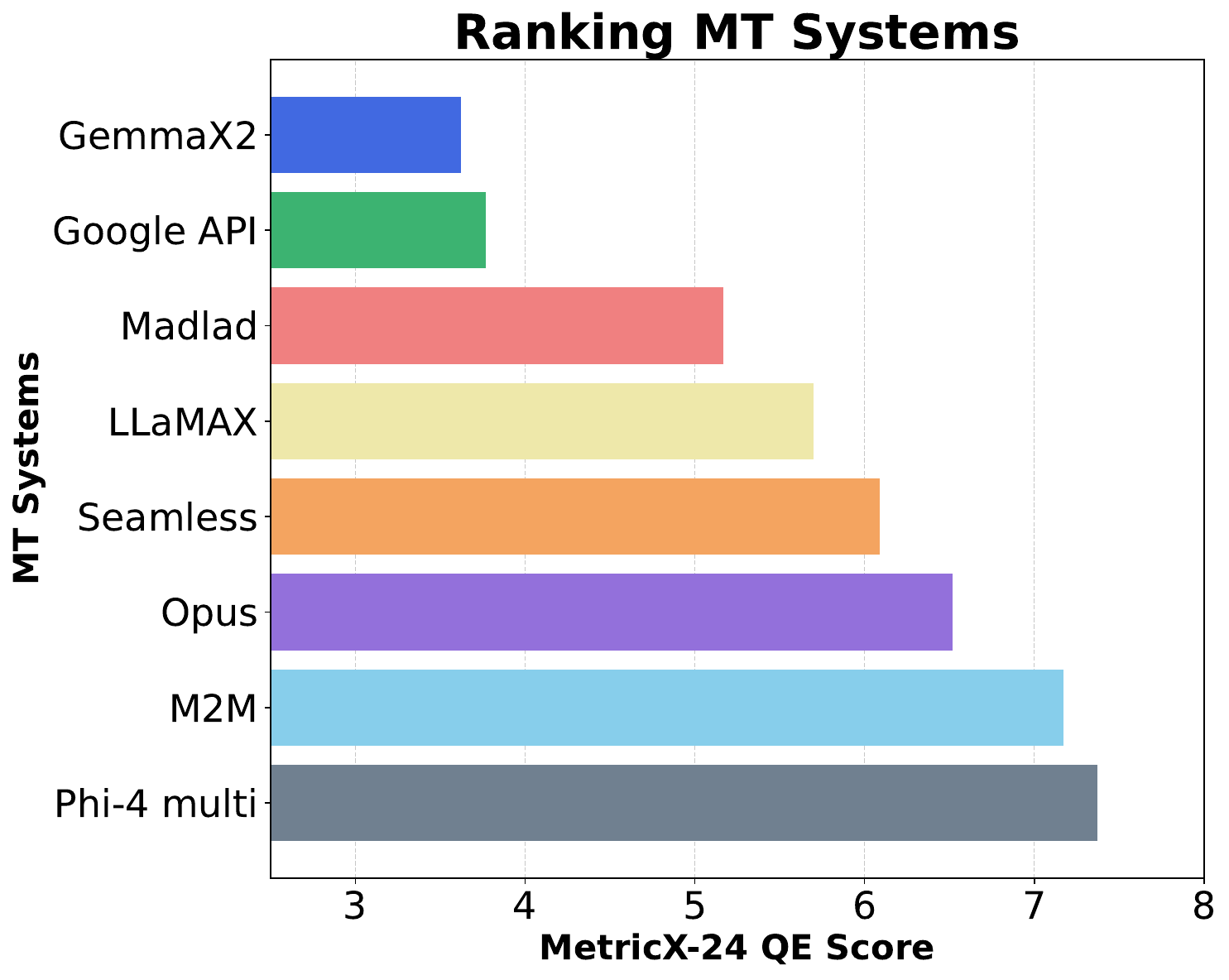}
    \caption{Ranking MT systems based on translation quality for sentences with VMWE using MetricX-24. Lower score indicates better performance.}
    \label{fig:system_rank_diagram}
\end{figure}

In this section, we rank translation quality across languages and MT systems using their MetricX-24 QE scores. xCOMET rankings appear in Appendix~\ref{app:xcomet_mt_rank}. For languages, we rank by QE scores averaged over all three VMWE categories and eight MT systems;  Figure~\ref{fig:language_rank_diagram} presents the results. Lower scores indicate better translation quality. The rankings reveal consistent cross-linguistic trends: high-resource languages such as German (de) and Spanish (es) perform best, whereas morphologically rich or low-resource languages such as Turkish (tr), Japanese (ja), and Czech (cs) perform worst. For MT systems, we rank by QE scores averaged over the three VMWE categories and seven languages; Figure~\ref{fig:system_rank_diagram} presents the results. GemmaX2 achieves the best translation quality, closely followed by Google API. Because our results exclude outputs in the wrong target language (Appendix~\ref{invalid_translations}) and GemmaX2 produces many such errors (Table~\ref{table:error_percentage}), its score is not directly comparable to systems with fewer failures. Phi-4-multi and M2M exhibit the largest degradation in presence of VMWEs. Interestingly, the best- and worst-performing systems are both LLM-based, while traditional neural MT systems occupy the middle ground, suggesting that scale alone does not guarantee VMWE robustness.

\section{Analyzing Translations With Low QE Scores (Error Analysis)}
\label{error_analysis}

\begin{CJK*}{UTF8}{gbsn}

\textbf{Example Error 1 - Sentence Fluency and Coherency}\\
Model: Google-API \\
Source Sentence: ``Once you're retired you have three months to get out.'' \\
Chinese Translation: ``一旦退休，您就有三个月的时间退休。'' \\
Explanation: The English sentence contains the verbal multiword expression ``get out'', which is often idiomatic. In this context, ``get out'' does not mean just to leave physically--it is often idiomatic for vacating a position, property, or situation. For the Chinese translated text, the MT system did not interpret ``get out'' idiomatically. In this context, it should mean something like ``vacate,'' ``leave the premises,'' or ``exit the position''. But it got mistranslated as another ``retire'', which loses the intended meaning. The original says ``Once you’re retired…'', but then the Chinese translation says ``三个月的时间退休''---i.e., ``three months to retire''---which repeats the idea of retiring, making the translated text something like ``Once you're retired, you have three months to retire.''\\
\\
\noindent \textbf{Example Error 2 - Literal Translation}\\
Model: LLaMAX3 Alpaca \\
Source Sentence: ``Young women were beginning to get on his nerves '' \\
Czech Translation: ``Young women začínaly být jeho nervy .'' \\
Explanation: This sentence contains the verbal multiword expression: ``get on his nerves''---expressing growing annoyance. The MT system literally translates the phrase to Czech---``začínaly být jeho nervy'' is grammatically broken and semantically incorrect, failing to convey the emotional content.\\
\\
\noindent \textbf{Example Error 3 - Literal Translation}\\
Model: GemmaX2 \\
Source Sentence: ``He 's dead to the world , ’ said Mabel , over her shoulder .'' \\
Chinese Translation: ``他已经死了，”梅布尔回头说道。'' \\
Explanation: This sentence contains the verbal multiword expression: ``dead to the world''---meaning to be deeply asleep, completely unaware of what's happening around. It does not mean actually dead. The translated sentence has a severe misinterpretation of the idiom---``他已经死了'' means ``He is already dead.''. This is a literal translation, completely losing the intended idiomatic meaning. The contextual mismatch means anyone reading this in Chinese would think a character died, changing the entire tone of the scene. Meanwhile, in the English version, it’s a light, perhaps even humorous or casual remark about someone being fast asleep.\\
\\
\noindent \textbf{Example Error 4 - Literal Translation}\\
Model: GemmaX2 \\
Source Sentence: ``Breaking up, as Neil Sedaka once put it, is hard to do.'' \\
Spanish Translation: ``Romper, como dijo una vez Neil Sedaka, es difícil de hacer.'' \\
Explanation: This sentence contains the verbal multiword expression: ``breaking up''---meaning to end a romantic relationship. It does not mean actually dead. ``Romper'' in Spanish means ``to break'', the Spanish sentence when back translated says ``To break, as Neil Sedaka once said, is difficult to do.'' losing its intended meaning.\\
\\
\noindent \textbf{Example Error 5 - Invalid Output (Mis-targeted Translation)}\\
Model: GemmaX2 \\
Source Sentence: ``At about the same time the aliens department of the Home Office took on extra staff and moved to Cleveland House on Thorney Street.'' \\
Turkish Translation: ``À peu près à la même époque, le département des étrangers du ministère de l'Intérieur a embauché du personnel supplémentaire et a déménagé à Cleveland House sur Thorney Street.'' \\
Explanation: Although the source sentence was intended to be translated into Turkish, the translation was mistakenly produced in French, demonstrating correct meaning transfer but in the wrong target language.\\
\\
\noindent \textbf{Example Error 6 - Invalid Output (Incomplete Translation)}\\
Model: Seamless \\
Source Sentence: ``Now the formula one grand prix circus is getting ready to hit the road for another season … first race is in South Africa in five weeks time … what 's new … is it going to be a head to head between Mansell and Senna.'' \\
Japanese Translation: ``フォーミュラ1グランプリ サーカス シーズン1のスタートを決めました'' \\
Explanation: In this instance, the verbal multiword expression ``hit the road'' means to get started. In this situation, the MT system does get the multiword expression correct, as the back-translation reads ``The Formula 1 Grand Prix circus has decided to start Season 1.'', however this is an example of incomplete translation as the MT system omitted the rest of the information in the source sentence. \\
\\
\noindent \textbf{Example Error 7 - Invalid Output (Incomplete Translation)}\\
Model: Google API \\
Source Sentence: ``So in effect, if you write this long, long, rambling press release, you won't get over the point you were trying to do, you'll get publicity okay, but you won't get the kind of thing you wanted.'' \\
Japanese Translation: ``長い長い'' \\
Explanation: In this instance, the Japanese text reads ``long long'', only translating a very small part of the sentence.\\
\\
\noindent \textbf{Example Error 8 - Invalid Output (No Translation)}\\
Model: LLaMAX3 Alpaca \\
Source Sentence: ``‘ Warm in winter and cool in summer, ’ Fen said.'' \\
German Translation: ``‘ Warm in winter and cool in summer, ’ Fen said.'' \\
Explanation: In this instance, the LLM outputs the English source sentence as its output without translating it.\\
\\
\noindent \textbf{Example Error 9 - Invalid Output (Partial Translation)}\\
Model: LLaMAX3 Alpaca \\
Source Sentence: ``Prior to becoming a Volunteer, I was a cutter/draper/pattern maker in a professional theatre.'' \\
Czech Translation: ``Prior to becoming a Volunteer, I was a řezník/sušíř/šikovník v profesionálním divadle.'' \\
Explanation: In this instance, the LLM partially translates the sentence.\\
\\
\noindent \textbf{Example Error 10 - Invalid Output (Character Repetition)}\\
Model: Google API \\
Source Sentence: ``I 'm , I 'm , quite amazed that, Poland was occupied by Germany and Czechoslovakia and France , that 's okay er and in fact you know we should just turn a blind eye to it and just let it carry on forever .'' \\
Japanese Translation: ``私は、この ...、この、この、この、この、この、この、、この、この、この、、この、この、この、、この、この、、この、この、、この、この、、この、この、、この、この、、この、この、、この、この、、この、....'' \\
Explanation: In this instance, the MT system repeats a character multiple times.

\end{CJK*}

\section{Error Span Overlap Details}
\label{ESO}

\begin{figure*}
    \centering
    \includegraphics[width=0.98\linewidth]{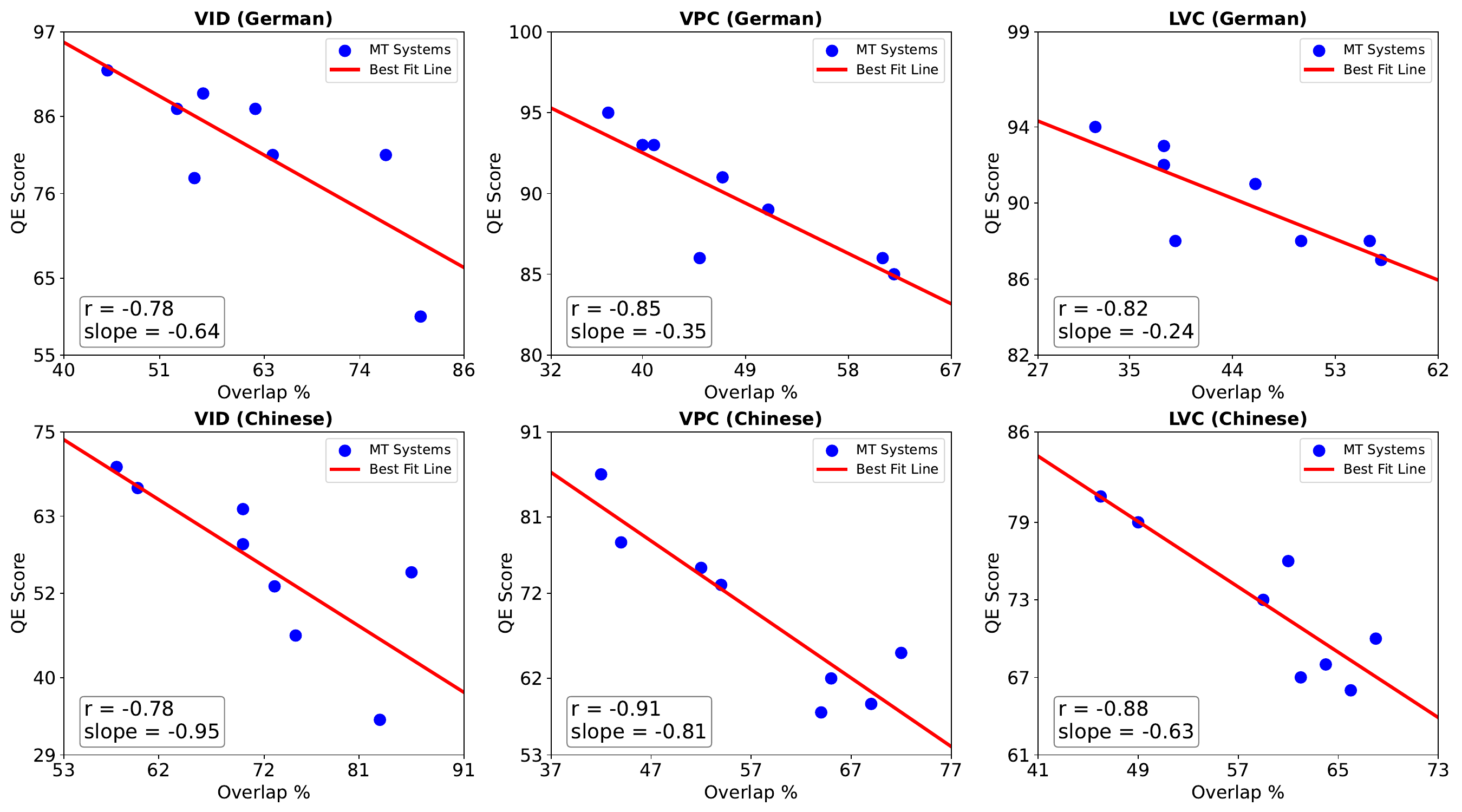}
    \caption{Error--span overlap vs.\ quality for German and Chinese. Each panel shows, for one VMWE type (VID, VPC, LVC), the relationship between the share of sentences whose xCOMET-QE error spans intersect the source VMWE (x-axis; lower is better) and the mean QE score (y-axis; higher is better). Blue points are MT systems; the red line is the least-squares fit. Insets report Pearson $r$ and the fitted slope. Across languages and VMWE types, greater overlap is strongly associated with lower quality (negative $r$), with the steepest effects for idioms.}
    \label{fig:correlation_graph}
\end{figure*}

We further analyze the xCOMET error-span overlap. Figure \ref{fig:correlation_graph} reports the Pearson correlation between the share of sentences where the error span overlaps the VMWE candidate and the overall QE score for Chinese and German. Strong correlations are found for all three VMWE categories, indicating that xCOMET spans covering the candidate are closely linked to lower translation quality rather than reflecting broader semantic or syntactic difficulty. Moreover, the fitted slope suggests that verbal idioms have the largest effect, followed by VPCs and then LVCs.

\section{Comparing Translation Quality across Languages \& MT Systems - xCOMET Result}
\label{app:xcomet_mt_rank}

\begin{figure}[t]
    \centering
    \includegraphics[width=0.98\linewidth]{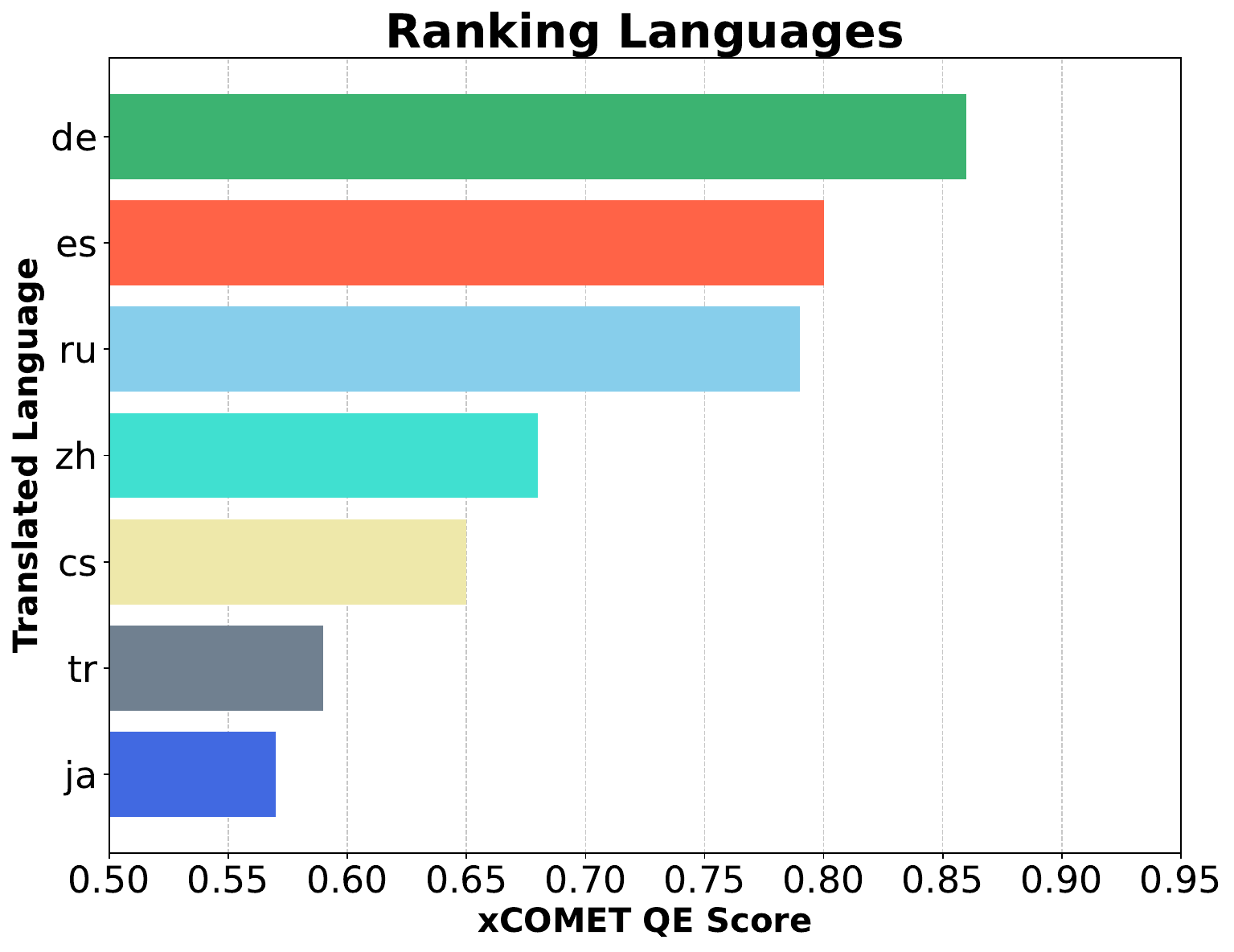}
    \caption{Ranking of languages based on the xCOMET QE scores.}
    \label{fig:xcomet_rank_lang}
\end{figure}

\begin{figure}[t]
    \centering
    \includegraphics[width=0.98\linewidth]{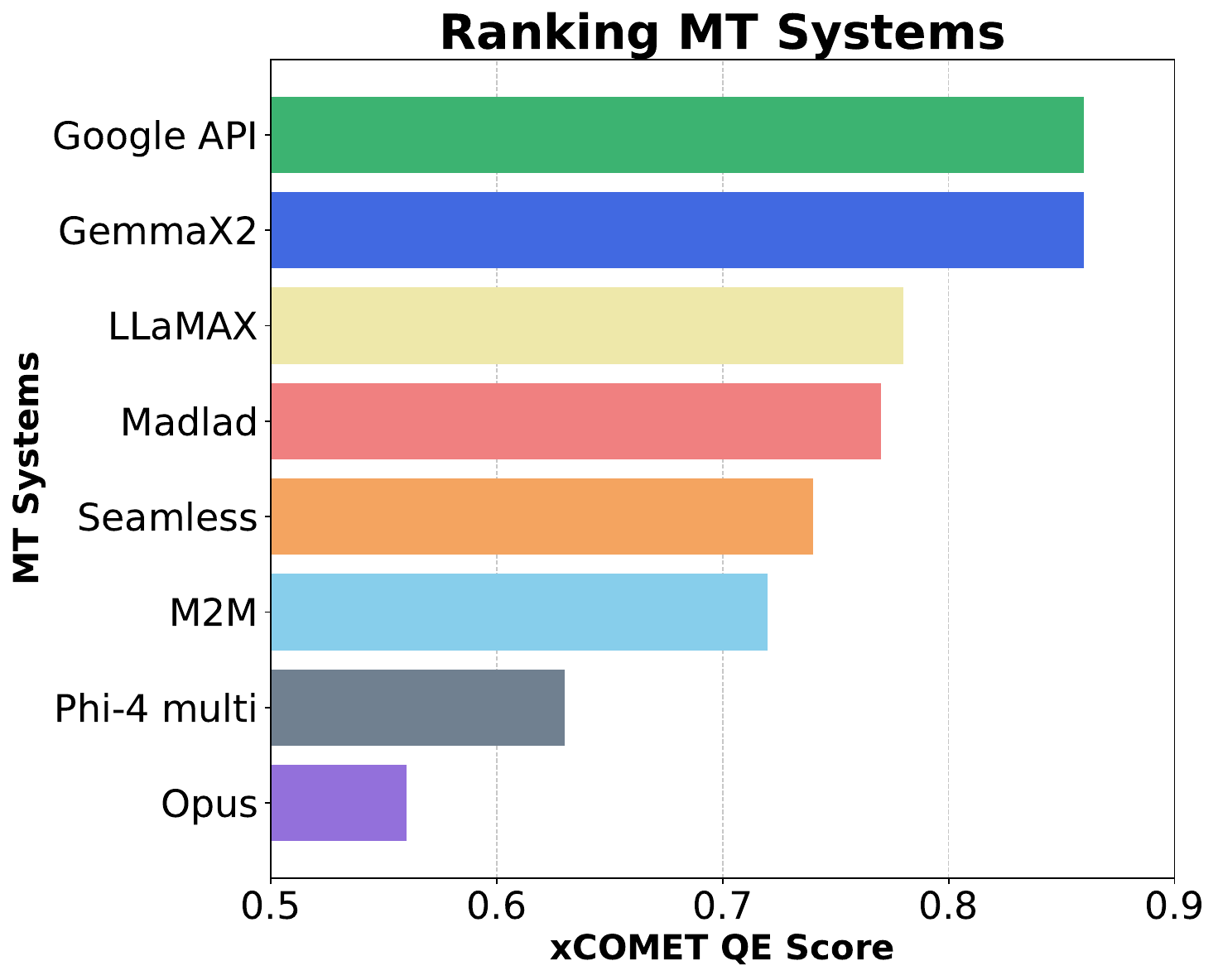}
    \caption{Ranking of MT systems based on the xCOMET QE scores.}
    \label{fig:xcomet_sys_lang}
\end{figure}

In this section, we rank translation quality across languages and MT systems. For languages, we rank them using their xCOMET QE scores averaged over all three VMWE categories and eight MT systems. Figure~\ref{fig:xcomet_rank_lang} shows the result. Higher scores indicate better translation quality. The rankings reveal consistent cross-linguistic trends: high-resource languages such as German (de) and Spanish (es) achieve the best performance, while morphologically rich or low-resource languages such as Turkish (tu), Japanese (ja), and Czech perform the worst. For MT systems, we rank them using the MetricX-24 QE scores averaged over three VMWE categories and seven languages. Figure~\ref{fig:xcomet_sys_lang} shows the result. Google API is the best performing model, closely followed by GemmaX2. In contrast, Phi-4-multi and M2M exhibit the largest degradation in presence of VMWEs.

\section{Prompt for GPT-based Translation}
\label{app:gpt_translation_prompt}

Table~\ref{tab:gpt_translation_prompt} reports the exact instruction prompt used for GPT-4.1 and GPT-5.1.

\begin{table*}[t]
    \centering
    \small
    \begin{tabular}{l p{0.75\linewidth}}
    \toprule
    \textbf{Model} & \textbf{Prompt} \\
    \midrule
    GPT-4.1 / GPT-5.1 &
    \begin{minipage}[t]{\linewidth}
    \ttfamily
    You are a professional machine translation system.
    Your job is to translate text from English into \{target\_lang\_name\}.
    For every input you receive, respond with ONLY the translated sentence in the target language.
    Do not include the source language, do not say ``Translation'', do not add quotes, explanation, or any extra text.
    \end{minipage}
    \\
    \bottomrule
    \end{tabular}
    \caption{Instruction prompt used to obtain GPT-4.1 and GPT-5.1 translations in our experiments.}
    \label{tab:gpt_translation_prompt}
\end{table*}

\section{Source Complexity and Possible Confounds}
\label{app:readability_confounds}

To check whether VMWE sentences are simply ``harder'' text, we computed Flesch--Kincaid readability scores for the evaluated sentences. This complements the confound controls already present in our regression analysis (e.g., length, lexical ambiguity proxies, and structural complexity).

\begin{table*}[t]
    \centering
    \begin{tabular}{lcl}
    \toprule
    \textbf{Corpus} & \textbf{Flesch Reading Ease} & \textbf{Evaluation} \\
    \midrule
    LVC & 51.40 & Fairly difficult (10th--12th grade) \\
    VPC & 64.83 & Standard (8th--9th grade) \\
    VID & 63.11 & Standard (8th--9th grade) \\
    non-VMWE & 56.53 & Fairly difficult (10th--12th grade) \\
    \bottomrule
    \end{tabular}
\caption{Readability statistics (lower Flesch score = harder to read).}
\label{tab:readability}
\end{table*}

Table~\ref{tab:readability} suggest that overall readability does not explain the observed translation degradation. The pattern actively contradicts a simple ``harder sentences $\rightarrow$ worse MT'' explanation. 

Specifically, VIDs and VPCs are more readable (higher Flesch scores) but show the largest translation degradation. Conversely, LVCs are less readable (lower Flesch score) but show the least translation degradation. These readability statistics help isolate the specific role of non-compositionality, reinforcing the regression-based confound analysis discussed in the main text.

\section{LLM Paraphrasing}

We evaluate previous proposed strategies for improving MT performance on VMWEs, particularly source side paraphrasing prior to translation~\citep{ donthi2025improvingllmabilitiesidiomatic, castaldo-monti-2024-prompting}. Our experiments show that while this approach proves effective for verbal idioms, it negatively affects the performance for light verbs, underscoring an inherent trade-off between overall generalization and improving specific VMWE categories.

LLM paraphrasing has been a tool proposed to help MT systems handle idioms~\citep{donthi2025improvingllmabilitiesidiomatic, castaldo-monti-2024-prompting}. Building on this line of work, we investigate if paraphrasing helps improve translation across different VMWE types. By replacing the VMWE span with a literal counterpart, we get a semantically similar sentence without the VMWE. This also quantifies how much of the observed degradation is due to the VMWE phrase itself, rather than the sentence’s semantic structure.

\begin{figure}[t]
    \centering
    \includegraphics[width=0.98\linewidth]{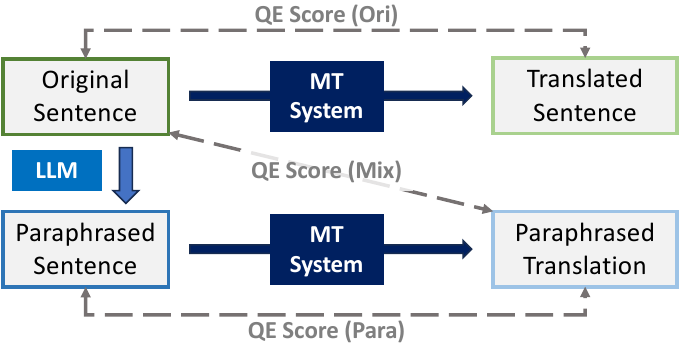}
    \caption{Paraphrasing structure. \textbf{Ori}: QE score between the original sentence and its direct translation. \textbf{Para}: QE score between paraphrased sentence and its direct translation. \textbf{Mix}: QE score between the original sentence and the translation of the paraphrased sentence.}
    \label{fig:paraphrase_diagram}
\end{figure}

We provide Llama-3.3-70B with the sentence and the VMWE phrase, and use a chain-of-thought prompting strategy to guide the model in replacing the verbal expression with a literal paraphrase, while preserving the overall semantic structure of the sentence. For example, the sentence containing an idiom ``\textit{He finally decided to \textbf{spill the beans}}'' is paraphrased to ``\textit{He finally decided to \textbf{reveal the secret}}.'' We then translate both original and paraphrased sentence to the target language. Given original sentence $s_o$, paraphrased sentence $s_p$, original sentence translation $t_o$, and paraphrased sentence translation $t_p$, we compute three QE scores as shown in Figure~\ref{fig:paraphrase_diagram}: Ori = $QE(s_o, t_{o})$, Mix = $QE(s_o, t_{p})$, and Para = $QE(s_p, t_{p})$. We report two derived scores using MetricX-24:\\
i) $\delta_{\text{mix}}$ = Ori - Mix. This score shows if paraphrasing the source before translation improves translation quality. A positive value means the paraphrasing improves the translation, negative value means the paraphrasing hurts the translation.\\ 
(ii) $\delta_{\text{para}}$ = Ori - Para. This score shows how much degradation in performance is attributed to the VMWE candidate, as we compare two semantically similar sentences with and without the VMWE candidate. Higher values mean the paraphrased sentence translates better than the VMWE-containing original, estimating how much the VMWE degrades translation (values near zero imply minimal VMWE effect).

Table~\ref{table:praphrasing_delta} reports the MetricX-24 results, full results for paraphrasing are given in Appendix \ref{Full_paraphrasing}. The scores are calculated as the average of all eight MT systems. The values for $\delta_{\text{para}}$ are positive (except Turkish LVC), indicating that the paraphrased non-VMWE sentence translates a lot better than the original VMWE counterpart. The values are highest in VID, followed by VPC, while LVC shows the least improvement. This trend is consistent across all languages. The results suggest that VIDs, which are the most non-compositional, suffer the most during translation. LVCs, which are the most compositional, are relatively unaffected.

The $\delta_{\text{mix}}$ values indicates if paraphrasing improves translation performance. Evidently, VID shows consistent improvement across all languages, demonstrating that paraphrasing is an effective strategy for this VMWE category. VPC also modestly benefits. However, LVC performance consistently declines. Overall, these findings reveal a key trade-off: \textit{paraphrasing enhances translation of highly non-compositional VMWEs but reduces domain-specific robustness, especially for compositional VMWE categories}. MT systems can leverage this strategy for VID and VPC cases to develop more robust, VMWE-aware translation systems.

\begin{table}
    \centering
    \resizebox{0.99\linewidth}{!}{
    \begin{tabular}{l|cc|cc|cc}
    \hline
    \multirow{2}{*}{\textbf{Lang}} & \multicolumn{2}{c|}{\bf VID} & \multicolumn{2}{c|}{\bf VPC} & \multicolumn{2}{c}{\bf LVC} \\
    \cmidrule(lr){2-3} \cmidrule(lr){4-5} \cmidrule(lr){6-7}
    & $\delta_{\text{mix}}$ & $\delta_{\text{para}}$ & $\delta_{\text{mix}}$ & $\delta_{\text{para}}$ & $\delta_{\text{mix}}$ & $\delta_{\text{para}}$ \\
    \hline
    Chinese & $+0.70$ & $+1.73$ & $+0.24$ & $+0.44$ & $-0.04$ & $+0.15$ \\
    German  & $+0.54$ & $+1.20$ & $+0.24$ & $+0.39$ &  $-0.17$ & $+0.10$ \\
    Russian & $+0.63$ & $+1.48$ & $+0.48$ & $+0.84$ & $-0.30$ & $+0.08$ \\
    Czech & $+1.06$ & $+2.02$ & $+0.51$ & $+0.86$ & $-0.29$ & $+0.20$ \\
    Spanish & $+0.68$ & $+1.49$ & $+0.38$ & $+0.68$ & $-0.27$ & $+0.10$ \\
    Japanese & $+0.12$ & $+0.61$ & $+0.12$ & $+0.30$  & $-0.17$ & $+0.06$  \\
    Turkish & $+0.37$ & $+1.46$ & $+0.33$ & $+0.77$  & $-0.53$ & $-0.07$ \\
    \hline
    \end{tabular}
    }
\caption{Paraphrasing results for MetricX-24 taking the average of all eight MT systems. $\delta_{\text{mix}}$: Difference between original and mix translation (Ori $-$ Mix). $\delta_{\text{para}}$: Difference between original and paraphrased translation (Ori $-$ Para). Positive $\delta$ value indicates paraphrasing improved the translation quality.}
\label{table:praphrasing_delta}
\end{table}

\subsection{Full Paraphrasing Results}
\label{Full_paraphrasing}

\begin{table*}[t]
    \centering
    \resizebox{0.99\linewidth}{!}{
    \begin{tabular}{p{0.6cm}l|ccc|ccc|ccc|ccc}
    \toprule
    \multirow{3}{*}{} & \multirow{3}{*}{\textbf{\makecell{MT\\System}}} & \multicolumn{3}{c}{\bf en-zh} & \multicolumn{3}{c}{\bf en-de} & \multicolumn{3}{c}{\bf en-ru} & \multicolumn{3}{c}{\bf en-cs} \\
    \cmidrule(lr){3-5} \cmidrule(lr){6-8} \cmidrule(lr){9-11} \cmidrule(lr){12-14}
    & & Ori & $\delta_{\text{mix}}$ & $\delta_{\text{para}}$ & Ori & $\delta_{\text{mix}}$ & $\delta_{\text{para}}$ & Ori & $\delta_{\text{mix}}$ & $\delta_{\text{para}}$ & Ori & $\delta_{\text{mix}}$ & $\delta_{\text{para}}$ \\
    \midrule
    \multirow{8}*{VID} & Madlad & 7.25 & $+1.77$ & $+2.21$ & 3.00 & $+0.24$ & $+0.92$ & 6.49 & $+0.87$ & $+1.76$ & 7.02 & $+0.24$ & $+1.23$ \\
     & Seamless & 6.98 & $+0.50$ & $+0.94$ & 3.72 & $+0.27$ & $+1.01$ & 6.72 & $+0.41$ & $+1.32$ & 8.65 & $+0.29$ & $+1.32$ \\
     & M2M & 7.07 & $+1.29$ & $+1.81$ & 5.53 & $+1.01$ & $+1.72$ & 9.56 & $+1.66$ & $+2.59$ & 11.63 & $+1.63$ & $+2.68$ \\
     & Opus & 7.95 & $+1.87$ & $+2.32$ & 6.87 & $+2.39$ & $+3.10$ & 9.50 & $+1.54$ & $+2.45$ & 13.41 & $+3.21$ & $+4.24$ \\
     & LLaMAX & 4.76 & $+0.24$ & $+0.72$ & 4.49 & $+0.27$ & $+0.90$ & 6.96 & $+0.38$ & $+1.21$ & 10.59 & $+0.58$ & $+1.48$ \\
     & Phi-4-multi & 4.63 & $+0.18$ & $+0.58$ & 4.22 & $+0.47$ & $+0.97$ & 7.34 & $+0.48$ & $+1.22$ & 14.39 & $+1.02$ & $+1.68$ \\
     & GemmaX2 & 3.31 & $-0.04$ & $+0.41$ & 2.41 & $-0.13$ & $+0.54$ & 4.69 & $-0.05$ & $+0.73$ & - & - & - \\
     & Google API & 2.89 & $-0.15$ & $+0.37$ & 1.92 & $-0.19$ & $+0.45$ & 3.69 & $-0.25$ & $+0.61$ & 7.88 & $+0.49$ & $+1.57$ \\
    \midrule
    \multirow{8}*{VPC} & Madlad & 6.46 & $+0.78$ & $+0.92$ & 2.44 & $+0.06$ & $+0.35$ & 5.72 & $+0.59$ & $+0.97$ & 6.71 & $+0.31$ & $+0.73$ \\
     & Seamless & 6.22 & $+0.20$ & $+0.39$ & 3.15 & $+0.15$ & $+0.44$ & 6.55 & $+0.59$ & $+0.96$ & 8.39 & $+0.65$ & $+1.12$ \\
     & M2M & 5.99 & $+0.45$ & $+0.71$ & 4.78 & $+0.70$ & $+1.01$ & 8.61 & $+1.15$ & $+1.53$ & 10.29 & $+1.16$ & $+1.65$ \\
     & Opus & 6.48 & $+0.65$ & $+0.90$ & 4.82 & $+0.74$ & $+1.07$ & 8.33 & $+0.87$ & $+1.27$ & 10.90 & $+1.10$ & $+1.58$ \\
     & LLaMAX & 4.40 & $+0.07$ & $+0.28$ & 3.87 & $+0.22$ & $+0.43$ & 6.64 & $+0.39$ & $+0.76$ & 9.64 & $+0.19$ & $+0.49$ \\
     & Phi-4-multi & 4.12 & $-0.08$ & $+0.07$ & 3.68 & $+0.11$ & $+0.33$ & 6.82 & $+0.23$ & $+0.59$ & 13.78 & $+0.28$ & $+0.50$ \\
     & GemmaX2 & 2.86 & $-0.12$ & $+0.07$ & 2.18 & $+0.07$ & $+0.32$ & 4.48 & $+0.18$ & $+0.43$ & - & - & - \\
     & Google API & 2.71 & $-0.02$ & $+0.20$ & 1.69 & $-0.07$ & $+0.16$ & 3.40 & $-0.11$ & $+0.21$ & 7.18 & $+0.42$ & $+0.86$ \\
    \midrule
    \multirow{8}*{LVC} & Madlad & 5.56 & $+0.37$ & $+0.59$ & 2.09 & $-0.19$ & $+0.09$ & 4.59 & $-0.17$ & $+0.21$ & 5.78 & $-0.39$ & $+0.12$ \\
     & Seamless & 6.00 & $-0.09$ & $+0.13$ & 2.70 & $-0.17$ & $+0.14$ & 5.41 & $-0.27$ & $+0.12$ & 7.00 & $-0.36$ & $+0.16$ \\
     & M2M & 5.24 & $-0.12$ & $+0.14$ & 3.69 & $-0.32$ & $+0.04$ & 6.80 & $-0.37$ & $+0.11$ & 8.54 & $-0.37$ & $+0.17$ \\
     & Opus & 5.71 & $-0.05$ & $+0.20$ & 3.95 & $+0.06$ & $+0.39$ & 6.90 & $-0.46$ & $-0.01$ & 9.49 & $-0.13$ & $+0.42$ \\
     & LLaMAX & 4.26 & $-0.17$ & $+0.04$ & 3.67 & $-0.14$ & $+0.15$ & 6.00 & $-0.25$ & $+0.14$ & 8.82 & $-0.32$ & $+0.16$ \\
     & Phi-4-multi & 4.28 & $-0.08$ & $+0.02$ & 3.14 & $-0.17$ & $+0.05$ & 6.06 & $-0.32$ & $+0.02$ & 12.47 & $-0.08$ & $+0.29$ \\
     & GemmaX2 & 2.94 & $-0.07$ & $+0.10$ & 1.82 & $-0.26$ & $-0.00$ & 3.88 & $-0.29$ & $+0.06$ & - & - & - \\
     & Google API & 2.65 & $-0.16$ & $+0.05$ & 1.39 & $-0.23$ & $+0.01$ & 3.05 & $-0.34$ & $+0.01$ & 6.09 & $-0.41$ & $+0.13$ \\
    \bottomrule
    \end{tabular}
    }
    \caption{Paraphrasing result. $\delta_{\text{mix}}$: Difference between original and mix translation (Ori $-$ Mix). $\delta_{\text{para}}$: Difference between original and paraphrased translation (Ori $-$ Para). Positive $\delta$ value indicates paraphrasing improved the translation quality.}
    \label{table:praphrasing_delta_detail}
\end{table*}

\begin{table*}
    \centering
    \resizebox{0.99\linewidth}{!}{
    \begin{tabular}{p{0.6cm}l|ccc|ccc|ccc|ccc}
    \toprule
    \multirow{3}{*}{} & \multirow{3}{*}{\textbf{\makecell{MT\\System}}} & \multicolumn{3}{c}{\bf en-zh} & \multicolumn{3}{c}{\bf en-de} & \multicolumn{3}{c}{\bf en-ru} & \multicolumn{3}{c}{\bf en-cs} \\
    \cmidrule(lr){3-5} \cmidrule(lr){6-8} \cmidrule(lr){9-11} \cmidrule(lr){12-14}
    & & Ori & $\delta_{\text{mix}}$ & $\delta_{\text{para}}$ & Ori & $\delta_{\text{mix}}$ & $\delta_{\text{para}}$ & Ori & $\delta_{\text{mix}}$ & $\delta_{\text{para}}$ & Ori & $\delta_{\text{mix}}$ & $\delta_{\text{para}}$ \\
    \midrule
    \multirow{8}*{VID} & Madlad & 35.55 & $+22.25$ & $+37.37$ & 89.15 & $-2.24$ & $+5.89$ & 77.14 & $+1.43$ & $+9.63$ & 68.15 & $-3.55$ & $+18.62$ \\
     & Seamless & 53.31 & $-2.76$ & $+16.72$ & 87.07 & $-2.36$ & $+6.01$ & 76.37 & $-0.21$ & $+8.51$ & 62.18 & $-4.01$ & $+18.94$ \\
     & M2M & 56.49 & $+0.08$ & $+17.05$ & 81.16 & $-0.57$ & $+8.81$ & 67.25 & $+3.48$ & $+12.47$ & 51.55 & $+1.15$ & $+23.40$ \\
     & Opus & 46.00 & $+2.91$ & $+23.56$ & 57.91 & $+22.93$ & $+31.96$ & 62.63 & $+6.73$ & $+15.79$ & 39.38 & $+10.90$ & $+33.96$ \\
     & LLaMAX & 66.10 & $-3.83$ & $+13.40$ & 81.70 & $+0.49$ & $+9.51$ & 73.78 & $+0.99$ & $+9.40$ & 51.33 & $-1.31$ & $+21.46$ \\
     & Phi-4-multi & 60.87 & $-2.88$ & $+15.86$ & 78.44 & $-2.04$ & $+11.19$ & 68.48 & $-0.06$ & $+11.84$ & 25.48 & $-0.31$ & $+13.60$ \\
     & GemmaX2 & 73.97 & $-4.15$ & $+11.32$ & 90.26 & $-2.33$ & $+5.40$ & 83.03 & $-1.08$ & $+6.12$ & - & - & - \\
     & Google API & 71.32 & $-1.42$ & $+15.61$ & 91.43 & $-2.52$ & $+5.09$ & 84.19 & $-0.45$ & $+7.58$ & 62.09 & $+0.63$ & $+22.57$ \\
    \midrule
    \multirow{8}*{VPC} & Madlad & 61.56 & $+4.29$ & $+8.79$ & 93.01 & $-0.16$ & $+2.02$ & 82.53 & $+1.81$ & $+4.11$ & 78.93 & $+0.00$ & $+6.94$ \\
     & Seamless & 64.49 & $+0.51$ & $+6.36$ & 91.15 & $-0.29$ & $+1.97$ & 79.87 & $+1.43$ & $+3.89$ & 72.61 & $+0.69$ & $+8.14$ \\
     & M2M & 66.83 & $+1.89$ & $+6.67$ & 85.66 & $+1.33$ & $+4.10$ & 73.92 & $+2.23$ & $+4.83$ & 65.80 & $+2.05$ & $+9.60$ \\
     & Opus & 60.99 & $+2.63$ & $+8.59$ & 86.40 & $+1.52$ & $+4.16$ & 73.72 & $+1.80$ & $+4.66$ & 63.03 & $+2.22$ & $+9.63$ \\
     & LLaMAX & 75.44 & $-0.53$ & $+4.00$ & 89.52 & $-0.17$ & $+2.32$ & 79.70 & $+0.53$ & $+2.92$ & 66.27 & $-0.96$ & $+5.58$ \\
     & Phi-4-multi & 73.64 & $-1.52$ & $+3.84$ & 86.85 & $-0.34$ & $+2.99$ & 76.96 & $+0.31$ & $+3.45$ & 34.41 & $-0.72$ & $+3.89$ \\
     & GemmaX2 & 82.62 & $-0.80$ & $+3.31$ & 93.81 & $-0.11$ & $+2.03$ & 86.33 & $+0.03$ & $+2.09$ & - & - & - \\
     & Google API & 84.26 & $-1.38$ & $+3.19$ & 95.26 & $-0.47$ & $+1.37$ & 89.91 & $-0.39$ & $+1.63$ & 76.37 & $+0.67$ & $+7.32$ \\
    \midrule
    \multirow{8}*{LVC} & Madlad & 70.56 & $+0.92$ & $+3.51$ & 94.62 & $-0.72$ & $+0.56$ & 86.65 & $-0.67$ & $+0.70$ & 85.63 & $-2.86$ & $+1.39$ \\
     & Seamless & 68.32 & $-0.84$ & $+2.05$ & 92.96 & $-0.71$ & $+0.51$ & 84.29 & $-0.84$ & $+0.75$ & 80.17 & $-1.85$ & $+2.41$ \\
     & M2M & 72.47 & $-1.49$ & $+1.29$ & 89.76 & $-0.59$ & $+0.72$ & 78.63 & $-0.59$ & $+1.09$ & 74.46 & $-1.83$ & $+2.50$ \\
     & Opus & 68.38 & $-1.76$ & $+1.56$ & 89.21 & $-0.06$ & $+1.27$ & 77.78 & $-1.04$ & $+0.73$ & 70.05 & $-1.17$ & $+3.53$ \\
     & LLaMAX & 78.04 & $-1.88$ & $+0.93$ & 90.56 & $-0.69$ & $+0.64$ & 81.30 & $-0.70$ & $+0.75$ & 72.22 & $-2.04$ & $+2.34$ \\
     & Phi-4-multi & 75.57 & $-1.37$ & $+1.21$ & 89.81 & $-1.23$ & $+0.51$ & 79.38 & $-1.12$ & $+0.53$ & 38.74 & $-1.14$ & $+1.62$ \\
     & GemmaX2 & 84.44 & $-1.28$ & $+1.10$ & 95.39 & $-0.85$ & $+0.11$ & 89.20 & $-0.75$ & $+0.26$ & - & - & - \\
     & Google API & 86.02 & $-1.83$ & $+0.53$ & 96.71 & $-0.92$ & $+0.23$ & 91.76 & $-0.77$ & $+0.56$ & 83.70 & $-2.54$ & $+1.75$ \\
    \bottomrule
    \end{tabular}
    }
\caption{Paraphrasing Result for xCOMET. All values are multiplied by 100 for better readability. $\delta_{\text{mix}}$: Difference between mixed translation and original translation (Mix - Ori). $\delta_{\text{para}}$: Difference between paraphrased translation and original translation (Para - Ori). Positive $\delta$ value indicates paraphrasing improved the translation quality.}
\label{table:praphrasing_delta_xcomet}
\end{table*}

Table \ref{table:praphrasing_delta_detail} and \ref{table:praphrasing_delta_xcomet} shows the full results of our paraphrasing experiment with MetricX and xCOMET as the evaluation model for the language pairs English to---Czech (\textit{en--cs}), German (\textit{en--de}), Russian (\textit{en--ru}) and Chinese (\textit{en--zh}). We don't report \textit{en-cs} scores for GemmaX2 due to the high translation errors. The observations align with table \ref{table:praphrasing_delta}. VID improves the most with the paraphrasing, followed by VPC and LVC. xCOMET seems to be more strict regarding the VID phrase paraphrasing, having mixed $\delta_{\text{mix}}$ values for all three categories, as compared to MetricX-24 which had mostly positive values in VID and VPC. Some MT systems gain substantially from paraphrasing, like Madlad (\textit{en-zh}) in VID gaining 22.22 and 37.31 in $\delta_{\text{mix}}$ and $\delta_{\text{para}}$.

Table \ref{table:metricx_other_language} presents the paraphrasing results for MetricX-24 for the language pairs English to---Spanish (\textit{en-es}), Japanese (\textit{en-ja}) and Turkish (\textit{en-tr}). Table \ref{table:xcomet_other_language} presents the results for xCOMET. Japanese has a very poor translation scores across both evaluation models for Opus. The trend for these three languages align with our primary observation, with MetricX-24 having generally favorable $\delta_{\text{mix}}$ scores, while xCOMET is more strict, having mixed results. $\delta_{\text{para}}$ improves the scores across both the evaluations, with the VID, VPC and LVC gaining the most in that order.

\begin{table*}
    \centering
    \resizebox{0.9\linewidth}{!}{
    \begin{tabular}{p{0.6cm}l|ccc|ccc|ccc}
    \toprule
    \multirow{3}{*}{} & \multirow{3}{*}{\textbf{\makecell{MT\\System}}} & \multicolumn{3}{c}{\bf en-es} & \multicolumn{3}{c}{\bf en-ja} & \multicolumn{3}{c}{\bf en-tr} \\
    \cmidrule(lr){3-5} \cmidrule(lr){6-8} \cmidrule(lr){9-11}
    & & Ori & $\delta_{\text{mix}}$ & $\delta_{\text{para}}$ & Ori & $\delta_{\text{mix}}$ & $\delta_{\text{para}}$ & Ori & $\delta_{\text{mix}}$ & $\delta_{\text{para}}$  \\
    \midrule
    \multirow{8}*{VID} & Madlad & 4.86 & $+0.44$ & $+1.29$ & 6.39 & $-0.27$ & $+0.34$ & 7.67 & $+0.05$ & $+1.24$ \\
     & Seamless & 5.35 & $+0.54$ & $+1.39$ & 8.17 & $+0.15$ & $+0.87$ & 8.53 & $+0.28$ & $+1.52$ \\
     & M2M & 7.76 & $+1.46$ & $+2.29$ & 8.02 & $+0.92$ & $+1.52$ & 11.93 & $+1.22$ & $+2.42$ \\
     & Opus & 7.22 & $+1.73$ & $+2.57$ & 21.11 & $+0.10$ & $-0.01$ & - & - & - \\
     & LLaMAX & 5.76 & $+0.43$ & $+1.18$ & - & - & - & - & - & - \\
     & Phi-4-multi & 5.61 & $+0.55$ & $+1.23$ & 6.78 & $+0.14$ & $+0.61$ & 14.23 & $+0.75$ & $+1.40$ \\
     & Google API & 3.38 & $-0.34$ & $+0.51$ & 4.39 & $-0.29$ & $+0.33$ & 5.69 & $-0.45$ & $+0.72$ \\
    \midrule
    \multirow{8}*{VPC} & Madlad & 4.19 & $+0.28$ & $+0.63$ & 6.40 & $+0.02$ & $+0.25$ & 7.11 & $+0.21$ & $+0.66$ \\
     & Seamless & 4.78 & $+0.41$ & $+0.75$ & 7.86 & $+0.32$ & $+0.56$ & 7.67 & $+0.18$ & $+0.67$ \\
     & M2M & 6.79 & $+1.09$ & $+1.47$ & 7.28 & $+0.50$ & $+0.76$ & 10.71 & $+0.78$ & $+1.28$ \\
     & Opus & 5.91 & $+0.97$ & $+1.29$ & 21.00 & $-0.07$ & $-0.10$ & - & - & - \\
     & LLaMAX & 5.08 & $+0.19$ & $+0.51$ & - & - & - & - & - & - \\
     & Phi-4-multi & 4.93 & $+0.18$ & $+0.49$ & 6.54 & $+0.10$ & $+0.21$ & 13.63 & $+0.48$ & $+0.78$ \\
     & Google API & 3.17 & $-0.05$ & $+0.29$ & 4.20 & $-0.11$ & $+0.15$ & 5.50 & $+0.04$ & $+0.50$ \\
    \midrule
    \multirow{8}*{LVC} & Madlad & 3.39 & $-0.31$ & $+0.07$ & 6.16 & $-0.26$ & $+0.01$ & 6.19 & $-0.55$ & $-0.06$ \\
     & Seamless & 3.87 & $-0.31$ & $+0.09$ & 7.51 & $-0.19$ & $+0.10$ & 6.90 & $-0.48$ & $+0.08$ \\
     & M2M & 5.12 & $-0.29$ & $+0.13$ & 6.63 & $-0.12$ & $+0.18$ & 8.95 & $-0.54$ & $-0.03$ \\
     & Opus & 4.60 & $-0.21$ & $+0.21$ & 22.27 & $-0.09$ & $+0.04$ & - & - & - \\
     & LLaMAX & 4.33 & $-0.32$ & $+0.01$ & - & - & - & - & - & - \\
     & Phi-4-multi & 4.26 & $-0.15$ & $+0.19$ & 6.12 & $-0.16$ & $+0.05$ & 11.61 & $-0.52$ & $-0.21$ \\
     & Google API & 2.69 & $-0.35$ & $+0.02$ & 4.10 & $-0.23$ & $+0.03$ & 4.71 & $-0.60$ & $-0.13$ \\
    \bottomrule
    \end{tabular}
    }
\caption{Paraphrasing Results for MetricX-24. Positive $\delta$ value indicates paraphrasing improved the translation quality.}
\label{table:metricx_other_language}
\end{table*}

\begin{table*}
    \centering
    \resizebox{0.9\linewidth}{!}{
    \begin{tabular}{p{0.6cm}l|ccc|ccc|ccc}
    \toprule
    \multirow{3}{*}{} & \multirow{3}{*}{\textbf{\makecell{MT\\System}}} & \multicolumn{3}{c}{\bf en-es} & \multicolumn{3}{c}{\bf en-ja} & \multicolumn{3}{c}{\bf en-tr} \\
    \cmidrule(lr){3-5} \cmidrule(lr){6-8} \cmidrule(lr){9-11} 
    & & Ori & $\delta_{\text{mix}}$ & $\delta_{\text{para}}$ & Ori & $\delta_{\text{mix}}$ & $\delta_{\text{para}}$ & Ori & $\delta_{\text{mix}}$ & $\delta_{\text{para}}$ \\
    \midrule
    \multirow{8}*{VID} & Madlad & 78.74 & $-1.71$ & $+11.82$ & 42.66 & $+6.06$ & $+32.80$ & 60.63 & $-2.28$ & $+20.38$ \\
     & Seamless & 78.33 & $-2.77$ & $+11.00$ & 39.92 & $-2.65$ & $+26.36$ & 59.56 & $-3.31$ & $+19.39$ \\
     & M2M & 69.26 & $+0.87$ & $+15.34$ & 52.41 & $-0.61$ & $+21.95$ & 47.39 & $-0.30$ & $+22.54$ \\
     & Opus & 53.99 & $+19.28$ & $+32.56$ & 18.32 & $+0.11$ & $-2.04$ & - & - & - \\
     & LLaMAX & 74.27 & $+0.08$ & $+13.54$ & - & - & - & - & - & - \\
     & Phi-4-multi & 64.07 & $-2.55$ & $+18.69$ & 52.38 & $-4.00$ & $+19.11$ & 27.28 & $-0.47$ & $+12.74$ \\
     & Google API & 84.61 & $-4.68$ & $+8.59$ & 76.05 & $-7.29$ & $+14.92$ & 71.86 & $-7.09$ & $+15.64$ \\
    \midrule
    \multirow{8}*{VPC} & Madlad & 86.40 & $+0.83$ & $+4.28$ & 68.79 & $-1.45$ & $+6.54$ & 73.70 & $-0.34$ & $+6.78$ \\
     & Seamless & 84.50 & $+1.18$ & $+4.83$ & 56.25 & $-0.32$ & $+7.69$ & 73.24 & $-0.60$ & $+6.15$ \\
     & M2M & 77.33 & $+3.48$ & $+7.47$ & 66.84 & $+0.37$ & $+7.16$ & 62.00 & $+0.72$ & $+7.88$ \\
     & Opus & 80.56 & $+2.78$ & $+6.69$ & 15.97 & $-0.19$ & $-0.07$ & - & - & - \\
     & LLaMAX & 83.98 & $+0.57$ & $+4.13$ & - & - & - & - & - & - \\
     & Phi-4-multi & 77.47 & $+0.63$ & $+5.95$ & 65.91 & $-1.11$ & $+5.12$ & 34.47 & $+0.03$ & $+4.82$ \\
     & Google API & 89.79 & $-0.19$ & $+3.21$ & 86.36 & $-1.77$ & $+4.42$ & 82.34 & $-1.45$ & $+4.66$ \\
    \midrule
    \multirow{8}*{LVC} & Madlad & 90.52 & $-1.40$ & $+0.75$ & 76.39 & $-3.34$ & $+1.26$ & 80.70 & $-2.88$ & $+0.73$ \\
     & Seamless & 89.32 & $-1.57$ & $+0.51$ & 63.97 & $-2.79$ & $+1.42$ & 78.54 & $-2.52$ & $+1.05$ \\
     & M2M & 84.43 & $-1.28$ & $+0.81$ & 73.13 & $-1.77$ & $+1.84$ & 69.81 & $-2.59$ & $+0.61$ \\
     & Opus & 85.87 & $-1.04$ & $+1.05$ & 16.52 & $-0.00$ & $-0.01$ & - & - & - \\
     & LLaMAX & 87.94 & $-1.36$ & $+0.45$ & - & - & - & - & - & - \\
     & Phi-4-multi & 83.51 & $-2.02$ & $+0.85$ & 70.53 & $-2.33$ & $+1.01$ & 42.98 & $-2.69$ & $+0.05$ \\
     & Google API & 93.04 & $-1.65$ & $+0.27$ & 89.96 & $-2.49$ & $+0.64$ & 88.17 & $-3.23$ & $+0.21$ \\
    \bottomrule
    \end{tabular}
    }
\caption{Paraphrasing Result for xCOMET. Positive $\delta$ value indicates paraphrasing improved the translation quality.}
\label{table:xcomet_other_language}
\end{table*}

\section{LLM Prompts for VMWE Extraction}
\label{app:llm-extract}

Table \ref{prompts} shows the prompts we used for GPT-4o, Llama-3.3-70B-Instruct, Phi-4-multi and DeepSeek-R1-Distill-Llama-70B classifying VMWE candidate sentences from the WMT dataset. The VMWE candidate sentences are collected as discussed in Appendix~\ref{app:llm_comparison_sec}

\section{LLM Prompts for VMWE Candidate Paraphrasing}

Table \ref{paraphrase_prompts} shows the prompts we used for paraphrasing the VMWE candidate. We use Llama 3.3 70B for this experiment, with temperature 0.9 and top-p value of 0.9. We use one or two few-shot examples in each category.

```latex
\begingroup
\renewcommand{\arraystretch}{1}
\onecolumn
{\small
\begin{longtable}{|p{0.1\linewidth}|p{0.8\linewidth}|}
\caption{Prompts for VMWE extraction from WMT dataset.}\label{prompts}\\

\hline
\bf VMWE & \bf Prompt \\
\hline
\endfirsthead

\multicolumn{2}{l}{\textit{(Continued from previous page)}}\\
\hline
\bf VMWE & \bf Prompt \\
\hline
\endhead

\multicolumn{2}{r}{\textit{(Continued on next page)}}\\
\endfoot

\hline
\endlastfoot

LVC &
Linguistic Analysis Task: Multi-choice Classification of Verb-Noun Constructions.

Context Sentence: {sentence}

Target Components:- VERB: {verb\_lemma} - NOUN: {noun\_lemma}

LVC-specific decision tree:

Apply test LVC.0 - [N-ABS: Is the noun abstract?]

No -> It is not an LVC, exit

Unsure -> Apply test LVC.1 - [N-PRED: Is the noun predicative?]

No -> It is not an LVC, exit

Yes/Unsure -> Apply test LVC.2 - [V-SUBJ-N-ARG: Is the subject of the verb a semantic argument of the noun?]

No -> It is not an LVC, exit

Yes/Unsure -> Apply test LVC.3 - [V-LIGHT: The verb only adds meaning expressed as morphological features?]

No -> It is not an LVC, exit

Yes -> Apply test LVC.4 - [V-REDUC: Can a verbless NP-reduction refer to the same event/state?]

No -> It is not an LVC, exit

Yes -> It is an LVC

Classification Options:

A: the noun is not abstract.

B: the noun is not predicative (i.e. lacks semantic arguments).

C: All tests pass – the construction qualifies as an LVC.

D: the verb’s subject is not a semantic argument of the noun.

E: the verb adds more meaning than mere morphological features.

F: a verbless NP cannot be formed referring to the same event/state.

Instructions: Provide your classification choice (A, B, C, D, E, or F) with brief reasoning in 3 sentences or less, ending with 'Final Answer: [Choice]'

Please provide your classification with reasoning as instructed.
\\
\hline

VPC &
Determine if the verb-particle combination is a VPC (phrasal verb) based on the following analysis.

Context: {sentence}

Combination: {verb\_lemma} {particle}

VPC Decision Tree:

1. Is the second element a particle (e.g., 'up', 'off')?

   - No → Not VPC (A)

   - Yes → Continue

2. Remove the particle from the combination. Does the remaining verb convey the same meaning as the full verb-particle phrase?

   - Yes → Not VPC (B)

   - No → Continue

3. Does the inclusion of the particle create a non-compositional meaning that is significantly different from the verb's original meaning?

   - No → Not VPC (C)

   - Yes → VPC (D)

A: Not a particle

B: Meaning remains similar without the particle

C: Particle does not significantly alter the meaning

"D: Valid VPC (Particle significantly alters meaning)

Answer with reasoning and 'Final Answer: [answer]

"Is this a valid VPC? Provide analysis.
\\
\hline

VID &
Verbal Idiom (VID) Classification Task

Context Sentence: {sentence}

Candidate Phrase: {candidate}

VID Decision Tree:

1. [CRAN] Contains cranberry word?

   Yes → VID

   No → Next test

2. [LEX] Regular replacement changes meaning?

   Yes → VID

   No → Next test

3. [MORPH] Morphological changes affect meaning?

   Yes → VID

   No → Next test

4. [MORPHSYNT] Morphosyntactic changes affect meaning?

   Yes → VID

   No → Next test

5. [SYNT] Syntactic changes affect meaning?

   Yes → VID

   No → Not VID

Examples:

- VID: 'kick the bucket', 'let the cat out of the bag'

- non-VID: 'take a walk', 'make a decision'

Instructions:

1. Analyze each test sequentially

2. Provide brief reasoning for each test

3. Conclude with 'Final Answer: [Yes/No]'

Is this candidate a Verbal Idiom (VID)? Apply the decision tree.
\\

\end{longtable}
}
\endgroup

\begingroup
\renewcommand{\arraystretch}{1}
\onecolumn
{\small
\begin{longtable}{|p{0.075\linewidth}|p{0.8\linewidth}|}
\caption{Prompts for VMWE candidate paraphrasing.}\label{paraphrase_prompts}\\

\hline
\bf VMWE & \bf Prompt \\
\hline
\endfirsthead

\multicolumn{2}{l}{\textit{(Continued from previous page)}}\\
\hline
\bf VMWE & \bf Prompt \\
\hline
\endhead

\multicolumn{2}{r}{\textit{(Continued on next page)}}\\
\endfoot
\endlastfoot

& \\
LVC & 
You are an expert in linguistics. Given a sentence containing a multi-word expression (VMWE), a Light Verb Construct (LVC). Your task is to rephrase the sentence to remove the VMWE while keeping the meaning intact. Think in the following steps.

Step 1 - Look at the phrase in the sentence that will be provided to you that constitutes the VMWE in the sentence.

Step 2 - Replace that phrase with an alternative wording that removes the VMWE while maintaining the sentence’s meaning.

Step 3 - Provide your output in the following format:

Rephrased Sentence: [Sentence without VMWE]

---Few-shot 1--

Sentence: She gave a smile before walking away. || Phrase: gave a smile

Step 1 - Provided VMWE phrase: The given phrase "gave a" is a light-verb construction (LVC), where "gave" (a light verb) combines with "smile" (a noun) to describe the act of smiling.

Step 2 - Replace the VMWE: A more lexicalized way to express this would be "smiled."

Step 3 - Output Format:

Rephrased Sentence: She smiled before walking away.

---Few-shot 2--
            
He took a deep breath before speaking. || Phrase: took a deep breath

Step 1 - Provided VMWE phrase: The given phrase "took a deep breath" is a light verb construction (LVC), where "take" (a light verb) combines with "breath" (a noun) to describe the act of inhaling.

Step 2 - Replace the VMWE: A more lexicalized way to express this would be "breathed deeply.

Step 3 - Output Format:

Rephrased Sentence: He breathed deeply before speaking.

Sentence: \textit{{sentence}} || Phrase: \textit{{candidate}} \\

& \\

\hline

&\\

VPC &   
You are an expert in linguistics. Given a sentence containing a multi-word expression (VMWE), a Verb Particle Construct (VPC). Your task is to rephrase the sentence to remove the VMWE while keeping the meaning intact. Think in the following steps.

Step 1 - Look at the phrase in the sentence that will be provided to you that constitutes the VMWE in the sentence.

Step 2 - Replace that phrase with an alternative wording that removes the VMWE while maintaining the sentence’s meaning.

Step 3 - Provide your output in the following format:

Rephrased Sentence: [Sentence without VMWE]

---Few-shot 1--

Sentence: His eyes welled up during the song. || Phrase: welled up

Step 1 - Provided VMWE phrase: The given phrase "welled up" is a verb-particle construction (VPC), where "well" (verb) combines with "up" (particle) to indicate the accumulation or rising of liquid, often referring to tears forming in the eyes.

Step 2 - Replace the VMWE: A more literal way to express this would be "filled with tears."

Step 3 - Output Format:

Rephrased Sentence: His eyes filled with tears during the song.

Sentence: \textit{{sentence}} || Phrase: \textit{{candidate}} \\

& \\

\hline

\vspace{0.3em}VID &
\vspace{0.3em}You are an expert in linguistics. Given a sentence containing a multi-word expression (VMWE), a Verbal Idiom (VID). Your task is to rephrase the sentence to remove the VMWE while keeping the meaning intact. Think in the following steps.

Step 1 - Look at the phrase in the sentence that will be provided to you that constitutes the VMWE in the sentence.

Step 2 - Replace that phrase with an alternative wording that removes the VMWE while maintaining the sentence’s meaning.

Step 3 - Provide your output in the following format:

Rephrased Sentence: [Sentence without VMWE]

---Few-shot 1--

Sentence: I think we still might take them up on it. || Phrase: take on [pron]

Step 1 - Provided VMWE phrase: The given phrase "take them up on" is a verbal idiom (VID), where "take up on" is an idiomatic expression meaning to accept an offer or invitation.

Step 2 - Replace the VMWE: A more literal way to express this would be "accept their offer."

Step 3 - Output Format:

Rephrased Sentence: I think we still might accept their offer.

---Few-shot 2--
  
Sentence: I did not pay much attention to it back then. || Phrase: pay to [pron]

Step 1 - Provided VMWE phrase: The given phrase "pay attention to" is a verbal idiom (VID), where "pay" (verb) combines with "attention" to mean focusing or noticing something.

Step 2 - Replace the VMWE: A more literal way to express this would be "notice."

Step 3 - Output Format:

Rephrased Sentence: I did not notice it much back then.
    
Sentence: \textit{{sentence}} || Phrase: \textit{{candidate}}\\

& \\

\hline

\end{longtable}
}
\endgroup

\end{document}